%% file: camera_ready.tex
\documentclass[sigconf]{acmart}

\AtBeginDocument{%
  \providecommand\BibTeX{{%
    \normalfont B\kern-0.5em{\scshape i\kern-0.25em b}\kern-0.8em\TeX}}}

\copyrightyear{2023} 
\acmYear{2023} 
\setcopyright{acmlicensed}\acmConference[KDD '23]{Proceedings of the 29th ACM SIGKDD Conference on Knowledge Discovery and Data Mining}{August 6--10, 2023}{Long Beach, CA, USA}
\acmBooktitle{Proceedings of the 29th ACM SIGKDD Conference on Knowledge Discovery and Data Mining (KDD '23), August 6--10, 2023, Long Beach, CA, USA}
\acmPrice{15.00}
\acmDOI{10.1145/3580305.3599543}
\acmISBN{979-8-4007-0103-0/23/08}

\graphicspath{{./figs/}}
\input{maths}

\usepackage{amsmath}
\usepackage{array}
\usepackage{multirow}
\usepackage{balance}
\usepackage{subcaption}
\usepackage{float}
\usepackage{graphicx}
\usepackage{bm}
\usepackage{CJKutf8}
\usepackage{multirow}
\usepackage{makecell}
\usepackage{booktabs}
\usepackage{dsfont}
\usepackage{dutchcal}
\usepackage{amsmath,lipsum}
\usepackage{cuted}
\usepackage{widetext}
\usepackage{xr}
\usepackage[utf8]{inputenc} 
\usepackage[T1]{fontenc}    
\usepackage{amsfonts}       
\usepackage{nicefrac}       
\usepackage{microtype}      
\usepackage{xcolor}         

\begin{document}

\title{Warpformer: A Multi-scale Modeling Approach for Irregular Clinical Time Series}

\author{Jiawen Zhang}
\authornote{This work was done during the internship at Microsoft Research Asia, Beijing, China.}
\affiliation{%
  \institution{The Hong Kong University of Science and Technology (Guangzhou)}
  \city{Guangzhou}
  \country{China}
}
\email{jzhang302@connect.hkust-gz.edu.cn}

\author{Shun Zheng}
\authornote{Corresponding Author}
\affiliation{%
  \institution{Microsoft Research Asia}
  \city{Beijing}
  \country{China}}
\email{shun.zheng@microsoft.com}

\author{Wei Cao}
\affiliation{%
  \institution{Microsoft Research Asia}
  \city{Beijing}
  \country{China}}

\email{wei.cao@microsoft.com}

\author{Jiang Bian}
\affiliation{%
 \institution{Microsoft Research Asia}
  \city{Beijing}
  \country{China}}
\email{jiang.bian@microsoft.com}

\author{Jia Li}
\authornotemark[2]
\affiliation{%
  \institution{The Hong Kong University of Science and Technology (Guangzhou)}
  \city{Guangzhou}
  \country{China}
  }
\email{jialee@ust.hk}

\renewcommand{\shortauthors}{Jiawen Zhang, Shun Zheng, Wei Cao, Jiang Bian, \& Jia Li}

\begin{abstract}
Irregularly sampled multivariate time series are ubiquitous in various fields, particularly in healthcare, and exhibit two key characteristics: intra-series irregularity and inter-series discrepancy.
Intra-series irregularity refers to the fact that time-series signals are often recorded at irregular intervals, while inter-series discrepancy refers to the significant variability in sampling rates among diverse series.
However, recent advances in irregular time series have primarily focused on addressing intra-series irregularity, overlooking the issue of inter-series discrepancy.
To bridge this gap, we present Warpformer, a novel approach that fully considers these two characteristics.
In a nutshell, Warpformer has several crucial designs, including a specific input representation that explicitly characterizes both intra-series irregularity and inter-series discrepancy, a warping module that adaptively unifies irregular time series in a given scale, and a customized attention module for representation learning.
Additionally, we stack multiple warping and attention modules to learn at different scales, producing multi-scale representations that balance coarse-grained and fine-grained signals for downstream tasks.
We conduct extensive experiments on widely used datasets and a new large-scale benchmark built from clinical databases.
The results demonstrate the superiority of Warpformer over existing state-of-the-art approaches.
\footnote{Our code is available at \url{https://github.com/imJiawen/Warpformer}}
\end{abstract}




\maketitle

\input{introduction}
\input{related_work}

\input{method}

\input{experiment}

\input{conclusion}


\bibliographystyle{ACM-Reference-Format}
\bibliography{main}
\balance

\newpage
\appendix

\input{supplementary}

\end{document}

%% file: maths.tex
\usepackage{amsmath,amsfonts,bm}
\usepackage{xfrac}

\def\eqref#1{equation~\ref{#1}}

\def\1{\bm{1}}

\DeclareMathAlphabet{\mathsfit}{\encodingdefault}{\sfdefault}{m}{sl}
\SetMathAlphabet{\mathsfit}{bold}{\encodingdefault}{\sfdefault}{bx}{n}

\newcommand{\R}{\mathbb{R}}



%% file: introduction.tex
\section{Introduction}
\label{sec:intro}

\begin{figure*}[t]
  \centering
    \subfloat[Six clinical time series]{
    \label{fig:exp1}
    \includegraphics[width=0.5\linewidth]{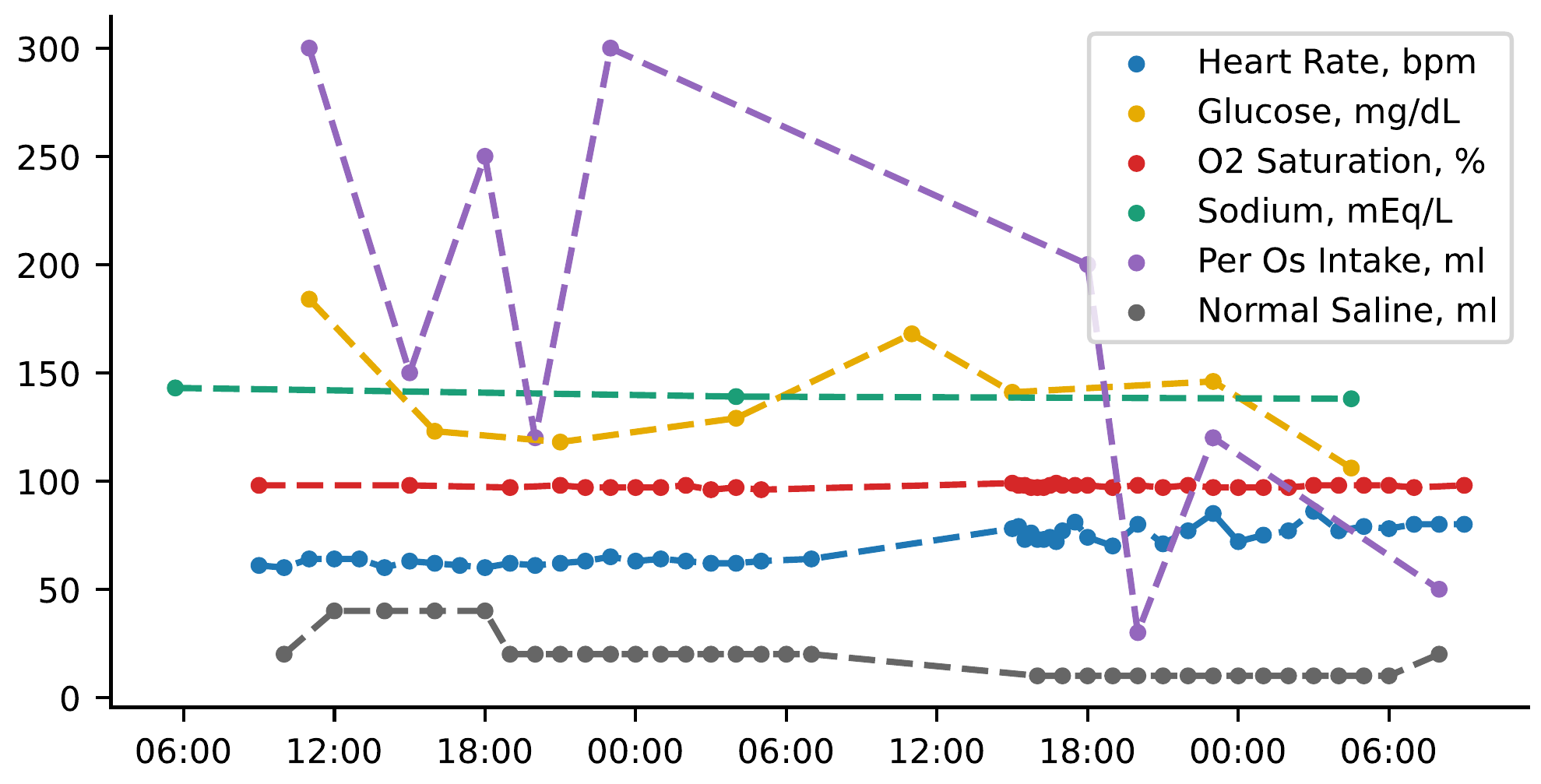}}
    \hspace{.3in}
    \subfloat[The distribution of sampling intervals for different clinical signals]{
    \label{fig:exp2}
    \includegraphics[width=0.4\linewidth]{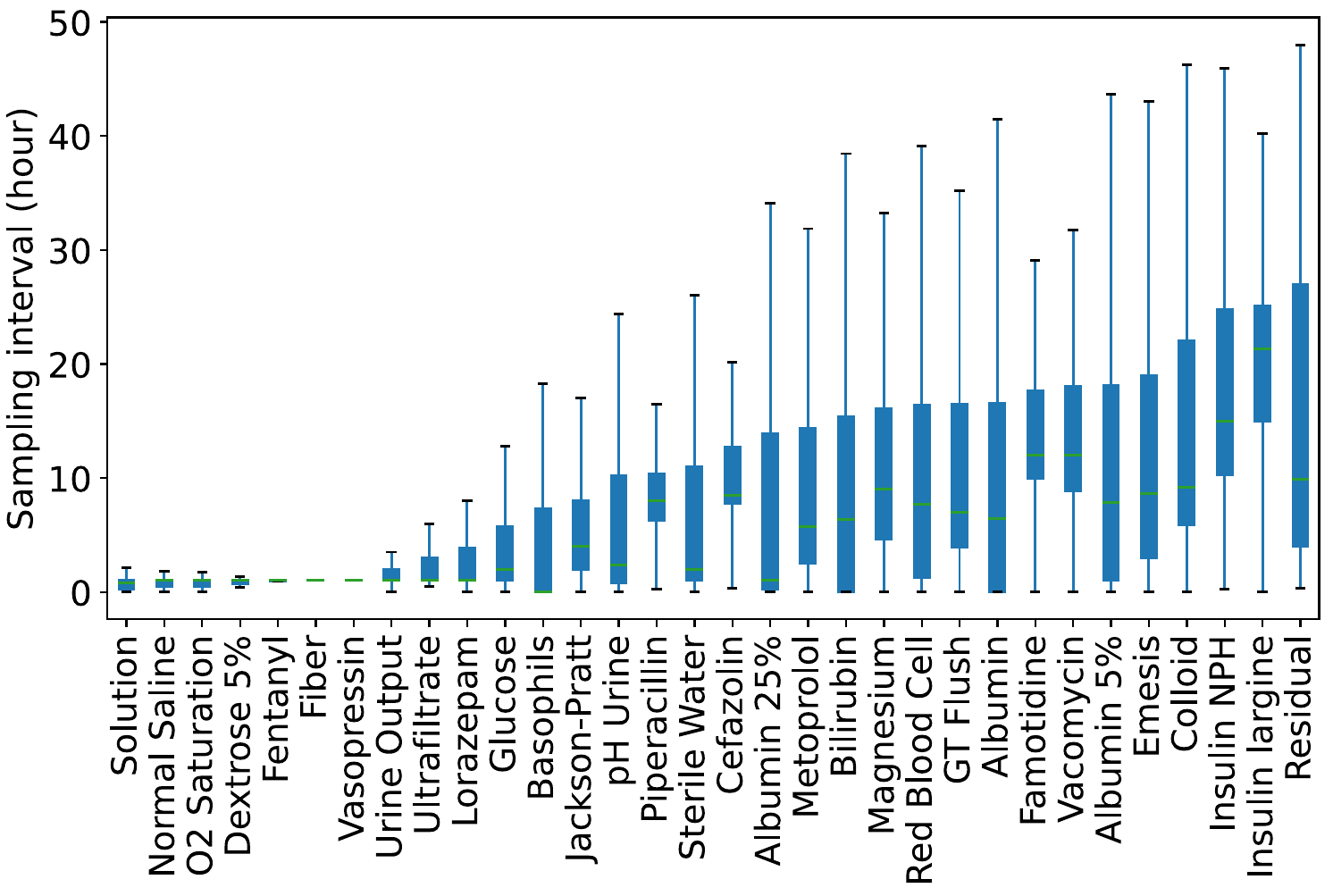}}
  \caption{\label{fig:examples} We select some representative time series from the MIMIC-III~\cite{johnson2016mimic} database to intuitively illustrate two prominent characteristics, \emph{intra-variate irregularity} and \emph{inter-variate discrepancy}, of irregularly sampled multivariate time series.}
\end{figure*}

With the rapid trend of digitalization in clinical systems, a large accumulation of clinical time-series data has attracted broad research attention, from both computer science and medical communities~\cite{jensen2012mining,yu2018artificial,liu2022human}, to explore machine learning solutions over diverse scenarios of digital therapeutics, such as risk stratification and early warning~\cite{zheng2020development, muralitharan2021machine, li2018tatc}, clinical outcome prediction~\cite{qian2021synctwin, heo2019machine}, and treatment recommendation~\cite{SymeonidisAZ21, nezhad2019deep}.
Advanced learning techniques over clinical time series indeed have profound impacts in the real world, such as improving the efficiency and the quality of clinical practices, relieving the burdens of clinicians and nurses, and in some sense, facilitating the equity in terms of the distribution of medical resources~\cite{johnson2016mimic,harutyunyan2019multitask,McDermott2021chil}.

As a specific type of irregularly sampled multivariate time series, clinical time series exhibit two prominent characteristics: \emph{intra-series irregularity} and \emph{inter-series discrepancy}.
\emph{Intra-series irregularity}refers to the fact that variate signals within each time series are usually recorded in irregular intervals in practice, breaking up the traditional assumption of regularly sampled time series.
Besides, \emph{inter-series discrepancy} denotes the presence of dramatically different sampling rates among multiple time series, leading to a significant imbalance in the occurrence of different signals.

To facilitate understanding, we select some representative time series from MIMIC-III~\cite{johnson2016mimic}, a large-scale healthcare database, and plot them in Figure~\ref{fig:exp1}.
We can observe that various vital physiological indicators, such as heart rate, glucose, oxygen saturation, and sodium, are irregularly measured, and some common clinical interventions, such as per os intake and normal saline, are also irregularly issued.
In addition to intra-series irregularity, we can also observe that heart rate is frequently measured while sodium is sporadically sampled, exhibiting significant discrepancy in sampling frequency.
Further, to provide a comprehensive view of inter-series discrepancy, we present the distribution of sampling intervals for tens of typical signals in Figure~\ref{fig:exp2}.

These two data characteristics present substantial challenges in modeling irregularly sampled multivariate time series, particularly in the context of clinical data.
First, \emph{intra-series irregularity} prevents the direct adoption of classical time-series models~\cite{zhou2021informer,Wu2021AutoformerDT,zhou2022fedformer,zhang2022lightts,liu2022pyraformer,shabani2023scaleformer,nie2023PatchTST} because these models do not include specific mechanisms to handle irregularity in time, which essentially conveys indispensable information on its own~\cite{che2018recurrent,zhang2021raindrop}.
Recent advances in irregular time series have made remarkable progresses to tackle \emph{intra-series irregularity}, such as extending recurrent neural networks~\cite{che2018recurrent}, leveraging neural ordinary differential equations (ODEs)~\cite{chen2018NeuralODE,Rubanova2019ODE-RNN}, and introducing time representations~\cite{Shukla2019IP-Net,Horn2020SeFunc,luo2020hitanet,li2019predicting, shukla2021multitime,zhang2021raindrop,tipirneni2022strats}.
Nevertheless, to the best of our knowledge, these methods do not pay attention to and address the dilemma caused by \emph{inter-series discrepancy} in unifying multivariate time series of significantly different granularities:
on one hand, a fine-grained unification retains detailed variations for frequently monitored signals but leads to very sparse placements of sporadically observed signals; on the other hand, a coarse-grained unification provides more balanced placements for all type of signals and can be used to obtain much clearer overall trending, but it sacrifices detailed variations for high-frequency signals.

To bridge this gap, this paper introduces Warpformer, a novel approach that fully considers both \emph{intra-series irregularity} and \emph{inter-series discrepancy}.
Warpformer starts with a specific input representation that unifies all signals in the original scale and encodes both signal values and other valuable information brought by \emph{intra-series irregularity}, such as sampling timepoints and intervals.
Besides, this representation preserves both temporal and signal dimensions to explicitly present the underlying inter-series discrepancy in sampling frequency.
Given this input representation, we stack several Warpformer layers, each of which consists of a warping module and a doubly self-attention module.
The warping module is a specifically designed differentiable network that directly operates on our representation for irregular time series and performs either down-sampling or up-sampling operations to unify all series in a new scale.
Following this, the doubly self-attention module, a Transformer~\cite{vaswani2017attention}-like network, is responsible for representation learning on the unified data representation.
As a result, multiple Warpformer layers together produce multi-scale representations for irregular time series, which are added through a residual connection to support downstream tasks.

We validate the effectiveness of Warpformer by comparing it with various state-of-the-art solutions on widely used datasets, such as PhysioNet~\cite{silva2012PhysioNet} and Human Activity~\cite{Rubanova2019ODE-RNN}.
Moreover, we also construct a new large-scale benchmark with five critical clinical tasks from the MIMIC-III~\cite{johnson2016mimic} database to provide more thorough comparisons.
Our results indicate that Warpformer, with its specific designs tailored to the characteristics of irregular time series, achieves significant improvements and becomes a new state-of-the-art on these benchmark datasets.

To sum up, our contributions include:
\begin{itemize}
    \item To the best of our knowledge, this work is the first one that highlights the importance of both \emph{intra-series irregularity} and \emph{inter-series discrepancy} for  irregularly sampled multivariate time series. Besides, Warpformer is also the first multi-scale approach for irregular time series.
    \item We provide extensive experiments to demonstrate the superiority of Warpformer and to verify our critical designs. More importantly, we have constructed a new benchmark with a much larger scale and diversified tasks, which can benefit future research in this field.
\end{itemize}

%% file: related_work.tex
\section{Related Work}
\label{sec:rel_work}

This work focuses on irregularly sampled clinical time series and highlights the multi-scale modeling capability.
Besides, we introduce a similar idea of dynamic time warping to enable the adaptive unification for irregular time series.
Thus we review related work from these aspects in this section.

\paragraph{\textbf{Modeling Irregularly Sampled Time Series}}
Existing methods for modeling irregularly sampled time series have apparent limitations.
Early methods~\cite{xu2018raim,MaGWZWRTGM20,MUFASA,ZhangQLLCGL21} used either hourly aggregation or forward imputation to obtain uniformly spaced intervals, overlooking meaningful temporal patterns in irregular sampling.
Later approaches, including modifying updating equations in recurrent neural networks~\cite{che2018recurrent,Mozer2017corr,Neil2016nips}, learning neural ODEs~\cite{chen2018NeuralODE,Rubanova2019ODE-RNN}, introducing time-based attentions~\cite{Shukla2019IP-Net,shukla2021multitime}, adding extra time representations~\cite{Horn2020SeFunc,luo2020hitanet,tipirneni2022strats}, and employing graphs to model interactions~\cite{zhang2021raindrop}, made remarkable progresses on capturing sampling irregularity but did not seriously consider significant discrepancy in sampling frequency across different series.
We note that two emerging paradigms from these approaches are more related to our work: 
one is to interpolate irregularly observed values to regularly spaced reference points~\cite{Shukla2019IP-Net,shukla2021multitime};
the other is to unfold irregular time series into a long sequence of "(value, type, time)" tuples~\cite{Horn2020SeFunc,luo2020hitanet,tipirneni2022strats}.
However, the former unifies all time series in one group of pre-specified reference time points, limiting its capability in identifying a data-oriented unification to balance fine-grained and coarse-grained information.
Besides, the latter is easy to suffer from the severe imbalance in the observations of different signals and lose the opportunity to capture the general trending hidden from the coarse-grained and balanced view.
In contrast, we seriously consider the challenge of \emph{inter-series discrepancy} and equip Warpformer with the multi-scale capability to balance fine-grained and coarse-grained views.

\paragraph{\textbf{Multi-scale Modeling for Time Series}}
It is well understood that multi-scale modeling plays a critical role in generic time series.
For instance, previous studies~\cite{ye2020lsan,MaGWZWRTGM20,luo2020hitanet} on electronic health records have shown the importance of capturing patient states with multiple time granularity.
Besides, recent studies~\cite{liu2022pyraformer,shabani2023scaleformer} have demonstrated that equipping Transformer extensions for time series with the multi-scale capability can bring significant performance improvements.
However, these multi-scale approaches only work in the context of regularly spaced time series, and it is non-trivial to adapt them into the setup of irregularly sampling.
Accordingly, Warpformer fills this gap and opens up an avenue for multi-scale analysis on irregularly sampled multivariate time series.

\paragraph{\textbf{Dynamic Time Warping}}

Existing studies~\cite{BrankovicBKNPW20,Sakoe1978DynamicPA,lohit2019temporal} across various application scenarios have witnessed the existence of temporal misalignment.
The underlying reasons could be phase shift, sampling rates, precision, etc.
To unify these misaligned signals, dynamic time warping (DTW) was proposed~\cite{berndt1994using} and further developed~\cite{cuturi2017soft, zhang2020time}.
Traditionally, people use DTW to adjust temporal matching by dynamic programming~\cite{BrankovicBKNPW20,Sakoe1978DynamicPA,berndt1994using,cuturi2017soft}.
In recent years, the need for input-dependent DTW has driven the development of various learning-based alignment solutions, such as temporal transformer nets~\cite{lohit2019temporal}, diffeomorphic temporal alignment nets~\cite{Weber2019DiffeomorphicTA}, dynamic temporal pooling~\cite{lee2021learnable}, and warping based on two-sided distributions~\cite{ScharwachterLM21}.
While sharing a similar idea of introducing warping when unifying different signals, this work fundamentally differs from existing DTW-based studies.
First, our warping module is the first design that connects differentiable DTW to irregular time series. In contrast, existing studies only consider DTW for regularly sampled time series.
Besides, we have formulated unique operations, which will be specified in Section~\ref{sec:method_warp}, to stimulate the adaptive unification of different signals in a given scale. This functionality is different from existing DTW studies that serve very different purposes, mostly for aligning and clustering.
Last, to the best of our knowledge, this is also the first time a warping module is embedded into a neural architecture to support multi-scale learning.

%% file: method.tex
\section{Methodology}
\label{sec:method}

\begin{figure*}[t]
  \centering
  \includegraphics[width=0.85\textwidth]{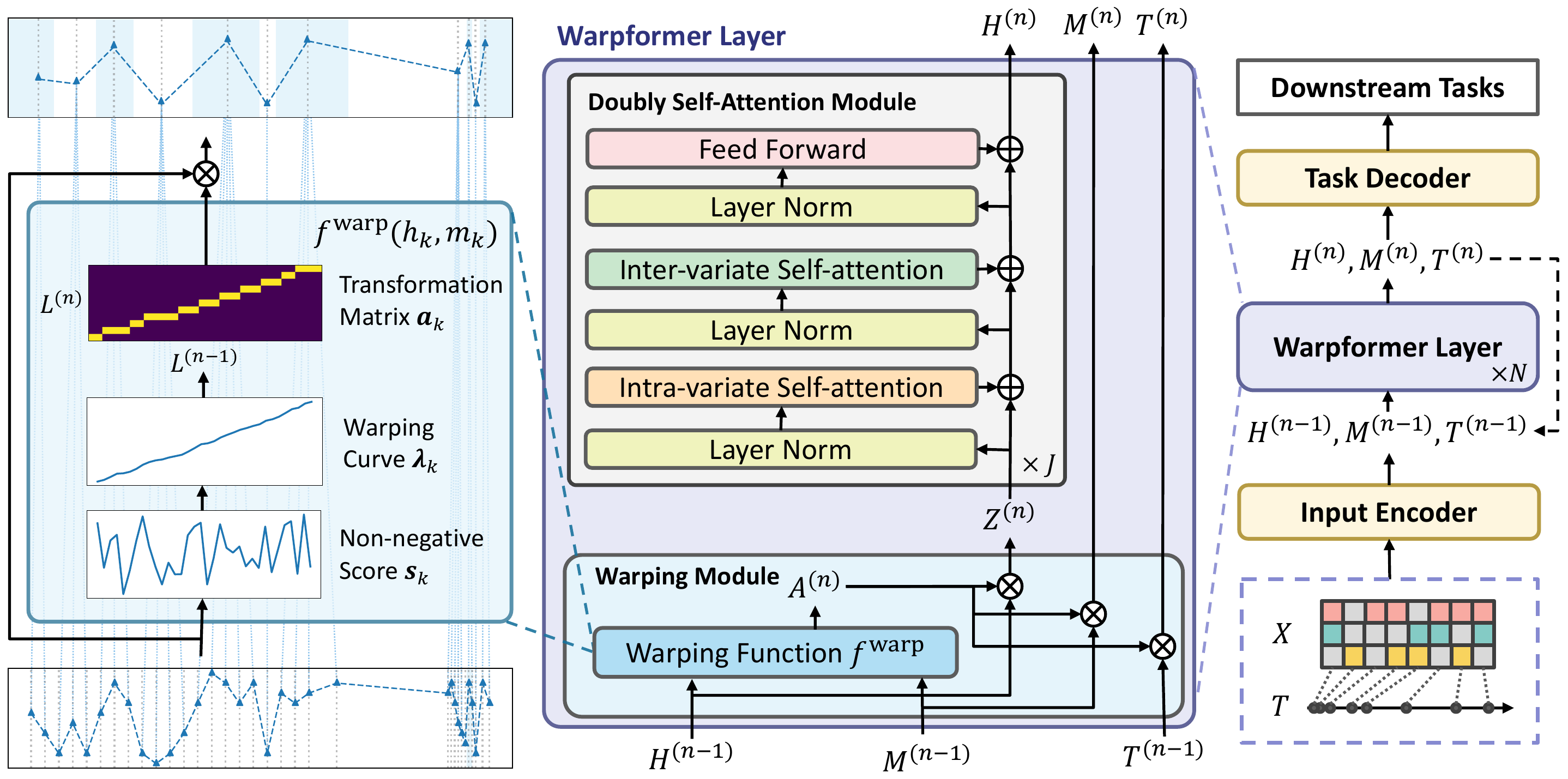}
  \caption{
  An overview of our Warpformer model:
  in the right side, we visualize the overall information flow in Warpformer;
  in the middle, we plot the detailed information flow within a Warpformer layer;
  in the left side, we visualize crucial intermediate results during the calculation of $f^{\textrm{warp}}$ for the $k$-th variate to facilitate understanding of how our warping module works.
  }
  \vspace{-0.1in}
  \label{fig:framework}
\end{figure*}

In Figure~\ref{fig:framework}, we give an overview of the information flow in Warpformer and highlights our specific designs within the Warpformer layer.
In a nutshell, Warpformer starts from a specific input encoder providing rich and structured data representations for irregular time series (Section~\ref{sec:method_input}) and stacks several Warpformer layers to enable multi-scale representation learning.
Each Warpformer layer consists of a warping module (Section~\ref{sec:method_warp}), unifying input representations in a given scale while in the meanwhile preserving the same structured format, and a doubly self-attention module (Section~\ref{sec:method_datt}), responsible for learning high-level representations from unified input data.
In this way, multiple Warpformer layers produce multi-scale representations, which are added via a residual connection and then fed into a task decoder (Section~\ref{sec:method_decoder}) to support downstream applications.

\subsection{Input Encoder}
\label{sec:method_input}

Let $\left\{[(t_i^k, x_i^k)]_{i=1}^{L^k}\right\}_{k=1}^K$ denote the irregularly sampled multivariate time series of a specific patient, where we have $K$ variates, the $k$-th variate contains $L^k$ irregularly sampled observations, and the $i$-th observation of the $k$-th variate is composed of the recording time $t_i^k$ and the recorded value $x_i^k$.
While holding the advantage of high flexibility in accommodating various signals with variable lengths and sampling rates, this data format does not fit naturally for batch processing in modern neural networks training and inference.
Thus, we turn to develop a specific data representation for Warpformer.

First, we collect all unique timestamps by taking an union operation over $[\{t_i^k\}_{i=1}^{L^k}]_{k=1}^K$ and organize them in ascending order as $T \in \R^L$, where $L$ denotes the number of all unique timestamps.
Then we fill in a value matrix $X \in \R^{K \times L}$, a type matrix $E \in \R^{K \times L}$, and a mask matrix $M \in \R^{K \times L}$ according to the following rules: $X_{k,j}=x_i^k, E_{k,j}=k, M_{k,j}=1$ if $T_j$ equals to $t_i^k$; $X_{k,j}=0, E_{k,j}=0, M_{k,j}=0$ otherwise.
Given such a data organization, we apply several encoding functions to obtain the input representation $H \in \R^{K \times L \times D}$ as $H = f^{\textrm{val}}(X) + f^{\textrm{type}}(E) + f^{\textrm{abs}}(T) + f^{\textrm{rel}}(T, M)$,
where $f^{\textrm{val}}$ denotes one fully-connected mapping to transform a single value to an embedding of size $D$, $f^{\textrm{type}}(\cdot)$ denotes a lookup table that transform categorical variate indicators into type embeddings (also size $D$), and $f^{\textrm{abs}}(\cdot), f^{\textrm{rel}}(\cdot, \cdot)$ are functions to obtain time-related embeddings.
Similar to encoding time-sensitive patterns in previous studies~\cite{shukla2021multitime,Wang2021OnPE}, we design $f^{\textrm{abs}}(\cdot)$, which is in essence a combination of several parallel sinusoidal mappings and one linear mapping, to encode the absolute time.
To be specific, the $d$-th element of $f^{\textrm{abs}}(\cdot)$'s output embedding is defined as $w_1 \cdot T + b_1$ if $d$ equals to $1$ and $\textrm{sin}(w_d \cdot T +b_d)$ otherwise ($1 < d \le D$).
Here $\{w_d, b_d\}_{d=1}^{D}$ make up the parameters of $f^{\textrm{abs}}(\cdot)$.
Besides, inspired by the success of relative positional encoding in sequence learning~\cite{Shaw2018SelfAttentionWR}, we develop $f^{\textrm{rel}}(\cdot, \cdot)$ to encode relative-time information.
In this function, we first calculate the sequence of sampling intervals for each variate based on $T$ and $M$ and then apply a two-layer perceptron over the interval value to obtain the relative-time embedding.
Last, we feed $H$ together with $M$ and $T$ (broadcasted from $\R^L$ to $ \R^{K \times L}$) to subsequent modules.

\paragraph{\textbf{Connection and Distinction Between Our Input Encoder and Existing Ones}}
The importance of incorporating distinctive information, such as sampling irregularity, into the learning process for irregularly sampled time series has been widely recognized~\cite{che2018recurrent,chen2018NeuralODE,shukla2021multitime,zhang2021raindrop}.
Sharing the same spirit, our input representation $H$ encapsulates various critical information, including not only variate values and types but also sampling time and frequencies.
But different from recent advancements that built on top of either regularly interpolated representations~\cite{Shukla2019IP-Net,shukla2021multitime} or a flattened sequence with varying types of signals interleaved~\cite{Horn2020SeFunc,luo2020hitanet,tipirneni2022strats}, our representation retains both temporal and variate dimensions ($\R^{K \times L}$) along with irregular intervals.
Such a structured format has several benefits, such as explicitly presenting concurrent observations of certain signals and, more importantly, clearly revealing inter-series discrepancies in sampling frequency.

\subsection{Warping Module}
\label{sec:method_warp}

The intention of introducing a warping module is to adaptively unify diverse irregularly sampled series into calibrated positions following a specific granularity for further representation learning.
Nevertheless, this functionality poses several requirements to the design of this module.
First, this module should produce input-dependent unification that fits for the underlying data distribution and benefits downstream tasks.
To this end, this module should be differentiable so that supervision signals from downstream tasks can guide appropriate unification.
Besides, this module should support both down sampling and up sampling to align both coarse-grained and fine-grained information.
Last, we want to preserve the warping output in the same format as the input $H \in \R^{K \times L \times D}$ owing to the same consideration of explicitly presenting \emph{inter-series discrepancy}, as specified in Section~\ref{sec:method_input}.
With a full consideration of these requirements, we develop our warping module as follows.

\paragraph{\textbf{An Overview.}}
Let ($H^{(n-1)}$, $M^{(n-1)}$, $T^{(n-1)}$) be the input to the warping module of the $n$-th Warpformer layer ($n \in \{1, \cdots, N\}$), and we use $L^{(n-1)}$ to denote the corresponding sequence length.
Our warping module aims to produce the transformation tensor $A^{(n)} \in \R^{K \times L^{(n)} \times L^{(n-1)}}$ based on the encoded representation $H^{(n-1)} \in \R^{K \times L^{(n-1)} \times D}$ and obtain the transformed representation $Z^{(n)} \in \R^{K \times L^{(n)} \times D}$, the new mask $M^{(n)} \in \R^{K \times L^{(n)} }$, and the new anchor time $T^{(n)} \in \R^{K \times L^{(n)}}$ as:
$A^{(n)} = f^{\textrm{warp}} \left(H^{(n-1)}, M^{(n-1)}\right)$,
$Z^{(n)} = A^{(n)} \otimes H^{(n-1)}$,
$M^{(n)} = A^{(n)} \otimes M^{(n-1)}$,
and $T^{(n)} = A^{(n)} \otimes T^{(n-1)}$,
where $f^{\textrm{warp}}(\cdot, \cdot)$ denotes a differentiable function that emits the warping transformation tensor $A^{(n)}$, $\otimes$ refers to a batch matrix product, and $L^{(n)}$ is a hyper-parameter pre-defined as the new sequence length in the $n$-th layer.
Note that, in order to support representation learning on the first layer over the raw granularity, we simply compose $K$ identity mappings into $A^{(1)}$ as $A^{(1)} = [I^{L \times L}_1,\cdots, I^{L \times L}_K]$, thus we have $H^{(1)} = H^{(0)} = H$, $M^{(1)} = M^{(0)} = M$, and $T^{(1)} = T^{(0)} = T$.
For subsequent layers that operate in adaptively identified granularity, we adopt a parameterized instantiation of $f^{\textrm{warp}}$.

\paragraph{\textbf{Calculating Warping Curves.}}
To facilitate understanding, below we specifically show how to obtain a mapping matrix $\bm{a}_k = A^{(n)}_{k, :, :} \in \R^{L^{(n)} \times L^{(n-1)}}$ from the enriched representation $\bm{h}_k = H^{(n-1)}_{k, :, :} \in \R^{L^{(n-1)} \times D}$ and the corresponding mask $\bm{m}_k = M^{(n-1)}_{k, :} \in \R^{L^{(n-1)}}$ of the $k$-th variate.
First, we compute a non-negative score for each observation entry in $\bm{h}_k$: $\bm{s}_k = f^{\bm{s}}(\bm{h}_k) \odot \bm{m}_k$,
where $\bm{s}_k \in \R^{L^{(n-1)}}$ is the score vector of the $k$-th variate, $f^{\bm{s}}(\cdot)$ is the score function, and $\odot$ denotes the element-wise product.
Since each entry in $\bm{h}_k$ already includes sufficient intra-series and inter-series information owing to the doubly self-attention module of the last layer, we simply instantiate $f^{\bm{s}}(\bm{h}_k)$ as a two-layer fully connected network.
In this way, we can obtain variate-specific scores and also take the global context into account.
Then we apply a normalized cumulative summation over $\bm{s}_k$ to obtain the warping curve $\bm{\lambda}_k \in \R^{L^{(n-1)}}$, so we have
$\bm{\lambda}_{k,i} = (\sum_{i'=1}^{i} \bm{s}_{k,i'}) / (\sum_{i'=1}^{L^{(n-1)}} \bm{s}_{k,i'})$.

\paragraph{\textbf{Calculating Transformation Mappings}}
This warping curve is non-descending, and its values are located within $[0, 1]$.
Therefore, by dividing $[0, 1]$ into $L^{(n)}$ segments and using segment boundaries to cut the warping curve $\bm{\lambda}_k$, we naturally derive a mapping from $\R^{L^{(n-1)}}$ to $\R^{L^{(n)}}$.
To be specific, we obtain the left boundaries of $L^{(n)}$ segments dividing $[0, 1]$ as $\bm{r}^1 = [0, \tfrac{1}{L^{(n)}}, \cdots, \tfrac{L^{(n)}-1}{L^{(n)}}]$ and the right boundaries as $\bm{r}^2 = [\tfrac{1}{L^{(n)}}, \tfrac{2}{L^{(n)}}, \cdots, 1]$.
To use $\bm{r}^1, \bm{r}^2 \in \R^{(n)}$ to cut $\bm{\lambda}_k \in \R^{(n-1)}$, we first extend the second dimension for $\bm{r}^1, \bm{r}^2$ and the first dimension for $\bm{\lambda}_k$ and then apply broadcast operations along the new dimension to expand them into the same space of $\R^{L^{(n)} \times L^{(n-1)}}$.
We denote the resulting matrices of $\bm{r}^1$, $\bm{r}^2$, and $\bm{\lambda}_k$ as $R^1$, $R^2$, and $\Lambda_k$, respectively.
Note that the operation of $\Lambda_k - R^1 \ge 0$ gives us the mask information indicating whether an element in $\lambda_k$ is on the right side of the left boundary of a specific segment.
Similarly, $R^2 - \Lambda_k \ge 0$ tells us whether an element is on the left side of the right boundary of a given segment.
Given the fact that an element belongs to a segment if and only if this element is located in the range covered by the left and right boundaries, we can obtain the matrix $a_k^m$ defining the mapping from $\R^{L^{(n-1)}}$ to $\R^{L^{(n)}}$ as
$\bm{a}_k^m = (\Lambda_k - R^1 \ge 0) \texttt{ \& } (R^2 - \Lambda_k \ge 0)$,
where \texttt{\&} denotes the logical And operation.

\paragraph{\textbf{Enabling Up-sampling.}}
Using the segment boundaries specified by $\bm{r}^1$ and $\bm{r}^2$ to cut the warping curve naturally supports the down-sampling behavior, which happens when several entries in $\bm{\lambda}_k$ fall into the same segment.
But this scheme cannot assign any entry in $\bm{\lambda}_k$ to multiple positions in $\R^{L^{(n)}}$ if we want a fine-grained unification when $L^{(n)} > L^{(n-1)}$.
To support up-sampling, our overall idea is to traverse the leftmost non-zero boundary in the upper triangle matrix $(\Lambda_k - R^1 \ge 0) \odot \Lambda_k$ and the rightmost non-zero boundary in the lower triangle $(R^2 - \Lambda_k \ge 0) \odot \Lambda_k$ and then copy these two boundary curves, essentially triggering up-sampling behaviors when $L^{(n)} > L^{(n-1)}$, to supplement the initial mapping matrix $\bm{a}_k^m$.
In practice, we can implement this idea very efficiently by adjusting $R^1$ and $R^2$.
First, we can use the values in the rightmost non-zero boundary of $(R^2 - \Lambda_k \ge 0) \odot \Lambda_k$ to update $\bm{r}^1$ as $\bm{r}^1_u = \min (\bm{r}^1, \textrm{perRowMax}((R^2 - \Lambda_k \ge 0) \odot \Lambda_k))$ and then broadcast $\bm{r}^1_u \in \R^{L^{(n)}}$ along a new dimension to obtain $R^1_u \in \R^{L^{(n)} \times L^{(n-1)}}$.
The adjusted $R^1_u$ ensures that the mappings recorded by the rightmost non-zero boundary of $(R^2 - \Lambda_k \ge 0)$ is definitely included in the non-zero entries of $(\Lambda_k - R^1_u \ge 0)$.
And we can perform a similar adjustment to obtain $R^2_u$.
Finally, we calculate $\bm{a}_k^m$ with new $R^1_u$ and $R^2_u$ as $\bm{a}_k^m = (\Lambda_k - R^1_u \ge 0) \texttt{ \& } (R^2_u - \Lambda_k \ge 0)$. 

\paragraph{\textbf{Enabling Differentiation.}}
Moreover, we note that the operation involved in obtaining $\bm{a}_k^m$ is non-differentiable. Accordingly, we utilize the following equation to compute a mapping matrix that preserves the flow of gradient computations:
$\bm{a}_k^u = \bm{a}_k^m \odot \left( \max \left(\Lambda_k - R^1, 0\right) + \max \left(R^2 - \Lambda_k, 0\right) \right)$.
Note that the value of a non-zero element in $\bm{a}_k^u$ denotes the summation of its distance to the left boundary and the distance to the right, which exactly equals to the segment width, namely $\frac{1}{L^{(n)}}$.
This fact means that $\bm{a}_k^u$ is a mapping matrix with uniform weights.
Furthermore, to stabilize the magnitude of the transformed embedding in $Z^{(n)}$, which should be irrelevant to the number of raw data points assigned to a specific segment, we normalize $\bm{a}_k^u$ by its per-row summation as $\bm{a}_k = \bm{a}_k^u  \big/ \text{rowSum}(\bm{a}_k^u)$.

It is apparent that the above operations to obtain $\bm{a}_k$ can be easily parallelized for all $K$ variates, and all these operations constitute the function $f^{\textrm{warp}}(H^{(n-1)}, M^{(n-1)})$ to produce the final transformation tensor $A^{(n)}$.
Moreover, we can observe that $T^{(n-1)}$ does not participate in the computation of $A^{(n)}$ because $H^{(n-1)}$ already includes time-related information in the input encoder.
We preserve the calculation from $T^{(n-1)}$ to $T^{(n)}$ mainly for visualization and interpretation purposes.

\subsection{Doubly Self-attention Module}
\label{sec:method_datt}

To perform effective learning over $Z^{(n)}$, the representation produced by the warping module in the $n$-th layer, a desired learning module on top of $Z^{(n)}$ should not only hold sufficient capacities to effectively capture predictive patterns but also retain the representation structure that includes both time and variate dimensions, in other words, keep operating in the space of $\R^{K \times L^{(n)} \times D}$, so as to preserve the inductive bias of explicitly presenting "concurrent" observations and to support further processing of subsequent warping modules.
Although Transformer~\cite{vaswani2017attention}, an extraordinarily successful neural architecture applied in various sequence learning scenarios~\cite{so2019evolved,dai2019transformer,dong2018speech,Li2018NeuralSS,BrownMRSKDNSSAA20}, appears to be a straightforward option, its self-attention mechanism, assuming the input sequence lies in the space of $\R^{L^{(n)} \times D}$, does not support our target of capturing both intra-series and inter-series patterns.
Therefore, we develop a novel doubly self-attention mechanism as a customized extension of the Transformer encoder to fit our representation structure.

Specifically, we perform two consecutive standard self-attention operations on two views of the organized representation $Z^{(n)} \in \R^{K \times L^{(n)} \times D}$.
In the first view, we regard $Z^{(n)}$ as $K$ sequences of embeddings with each sequence in $\R^{L^{(n)} \times D}$.
As for the second, we rearrange $Z^{(n)}$ into $L^{(n)}$ sequences of embeddings with each one in $\R^{K \times D}$.
Besides, each self-attention operation also follows prior arts~\cite{Wang2019LearningDT} to perform layer normalization~\cite{Ba2016LayerN} over the input in advance and include a residual connection.
Such a doubly self-attention mechanism enables the information exchange across both intra-series and inter-series dimensions.
After that, we finish one-layer encoding by applying the vanilla position-wise feed-forward mapping in Transformer, which naturally fits our representation format.
Last, to enable effective learning and facilitate sufficient exchanges of both intra-series and inter-series information, we stack multiple such encoding layers to compose the doubly self-attention module that maps $Z^{(n)}$ into $H^{(n)}$ for the subsequent process.
Note that $Z^{(n)}$ and $H^{(n)}$ are in the same space as $\R^{K \times L^{(n)} \times D}$, and we do not need to change $M^{(n)}$ and $T^{(n)}$.

\subsection{Task Decoder}
\label{sec:method_decoder}

With multi-layer stacking of Warpformer layers, each of which consists of a warping module and a doubly self-attention module, we can naturally obtain multi-scale representations by aggregating the output of each layer, denoted as $\{(H^{(n)}, M^{(n)})\}_{n=1}^N$.
While these representations correspond to different sequence lengths, denoted as $\{L^{(n)}\}_{n=1}^N$, we adopt attention-based pooling operations to obtain fix-sized embeddings for downstream tasks.
Note that these representations correspond to different sequence lengths, denoted as $\{L^{(n)}\}_{n=1}^N$, due to the layer-by-layer warping transformations towards more coarse-grained views.
To leverage these multi-scale representations in variable shapes, we condense the time and variate dimensions of each $H^{(n)}$ to obtain a fixed-size embedding in $\R^D$.
To be specific, we leverage an attention-based aggregation operation:
$f^{\textrm{agg}}(\textrm{Query}, \textrm{Key}, \textrm{Value}) = \text{Softmax}(\textrm{Query} \cdot \textrm{Key}^T) \cdot \textrm{Value}$,
where $\textrm{Query} \in \R^{D}$ is a query vector, $\textrm{Key} \in \R^{L \times D}$ is a key matrix, $\textrm{Value} \in \R^{L \times D}$ is a value matrix, and we use $\cdot$ to denote vector-matrix product.
Note that no matter how $L$ varies, the output of $f^{\textrm{agg}}$ is still in $\R^{D}$.
To condense the time dimension, we simply set $\textrm{Query} = Q^t$, $\textrm{Key} = \texttt{Tanh}(\texttt{Linear}(\bm{h}_k))$, $\textrm{Value} = \bm{h}_k$, $\bm{u}_k = f^{\textrm{agg}}(\cdot, \cdot, \cdot)$, where $Q^t \in \R^D$ denotes a parameter, $\bm{h}_k \in \left\{H^{(n)}_{k,:,:} \in \R^{L^{(n)} \times D} \right\}_{k=1}^K$ refers to the $k$-th slicing of $H^{(n)}$ along the variate dimension, and similarly $\bm{u}_k \in \R^D$ denotes the $k$-th slicing of $U^{(n)} \in \R^{K \times D}$.
Furthermore, we set $\textrm{Query} = Q^v$, $\textrm{Key} = \texttt{Tanh}(\texttt{Linear}(U^{(n)}))$, $\textrm{Value} = U^{(n)}$, and $\bm{v}^{(n)} = f^{\textrm{agg}}(\cdot, \cdot, \cdot)$ to remove the variate dimension.
In this way, we transform the tensor $H^{(n)}$ into the fixed-size embedding $\bm{v}^{(n)} \in \R^{D}$.
By applying the same operations with shared parameters over all $\{H^{(n)}\}_{n=1}^N$, we obtain a bunch of equal-size embeddings $\{\bm{v}^{(n)}\}_{n=1}^N$ encapsulating multi-scale patterns.
We add these embeddings into the final representaiton $\bm{v} = \sum_{n=1}^N \bm{v}^{(n)}$ to support downstream tasks.
For example, given a multi-class classification task with $C$ classes in total, we can perform the following computation:
$\hat{\bm{y}} = \texttt{Softmax}( W^y \bm{v} + \bm{b}^y)$,
where the prediction vector $\hat{\bm{y}} \in \R^{C}$ includes a predicted probability distribution over $C$ classes, and $W^y \in \R^{C \times D}, \bm{b}^y \in \R^C$ are task-specific parameters.

%% file: experiment.tex
\section{Experiments}
\label{sec:exp}

In this section, we provide extensive experiments to demonstrate the effectiveness of Warpformer.

\subsection{Experimental Settings}
\label{sec:exp_setting}

\subsubsection{\textbf{Datasets}}
We follow existing studies~\cite{che2018recurrent,chen2018NeuralODE,shukla2021multitime} to compare different models on the \emph{PhysioNet} dataset~\cite{silva2012PhysioNet} and the \emph{Human Activity} dataset~\cite{Rubanova2019ODE-RNN}.
The \emph{PhysioNet} dataset comprises $4,000$ instances and focuses on predicting in-hospital mortality. It includes $4$ types of demographics and $37$ physiological signals collected during the initial 48 hours of ICU admission. The median length of instances in this dataset is $72$.
The \emph{Human Activity} dataset aims to classify specific human activities among seven types for each timepoint in a segment of irregularly sampled time series. It consists of $6,554$ time-series segments with a total of $12$ channels. All instances in this dataset have a fixed length of $50$ timepoints.
We mainly follow the existing setup~\cite{shukla2021multitime} to divide the original dataset into the train, validation, and test sets, except that we do not shuffle instances in the \emph{Human Activity} dataset, which are obtained by truncating five long sequences, to avoid the potential information leakage.
Nevertheless, these two datasets cover very limited clinical prediction tasks domain, signal types, and data instances, which may lead to biased estimates in comparing different approaches.
Meanwhile, we note that the \emph{MIMIC-III} database~\cite{johnson2016mimic} contains many clinical scenarios calling for accurate predictions over diversified clinical signals, most of which are irregularly sampled.
Then we refer to~\cite{MIMIC-Extract, McDermott2021chil, tipirneni2022strats} and construct a new benchmark with five representative clinical tasks, 103 clinical signals (61 biomarkers and 42 interventions), and hundreds of thousands of instances.
The specific clinical tasks include in-hospital \underline{mor}tality (MOR), \underline{dec}ompensation (DEC), \underline{l}ength \underline{o}f \underline{s}tay (LOS), next timepoint \underline{w}ill \underline{b}e \underline{m}easured (WBM), and \underline{c}linical \underline{i}ntervention \underline{p}rediction (CIP), each of which serves as a new dataset for evaluation.
Additional details on these datasets can be found in Appendix~\ref{app:data:mimic3}.


\subsubsection{\textbf{Metrics}}
In line with previous studies, we evaluate the performance using the area under the receiver operating characteristic curve (AUROC) and the area under the precision-recall curve (AUPRC) for the \emph{PhysioNet} dataset and the five datasets derived from \emph{MIMIC-III}. Besides, following the prior art, we introduce Accuracy as the evaluation metric for the \emph{Human Activity} dataset.
For multi-class or multi-label classifications, we calculate AUROC and AUPRC scores for each individual class (label) and then compute the average as the dataset-level score. To eliminate the randomness, we conduct each experiment with five random seeds and report both the mean and standard deviation of the results.

\subsubsection{\textbf{Baselines}}
We organize existing methods applicable to irregular clinical time series into five paradigms.
The first is to introduce irregularity-sensitive updating mechanisms into recurrent neural networks, such as \textbf{RNN-Mean}, \textbf{RNN-Forward}, \textbf{RNN-$\Delta_t$}, \textbf{RNN-Decay}, and \textbf{GRU-D} in~\cite{che2018recurrent} as well as \textbf{Phased-LSTM}~\cite{Neil2016nips}.
The second is to model the hidden continuous dynamics behind irregular observations via neural ODEs, such as \textbf{ODE-RNN} and \textbf{L-ODE-ODE}~\cite{Rubanova2019ODE-RNN}.
Since neural ODEs are computational intensive, the third paradigm, including \textbf{IP-Net}~\cite{Shukla2019IP-Net} and \textbf{mTAND}~\cite{shukla2021multitime}, attempts to introduce continuous-time representations and align irregular observations into regularly spaced reference points to fit for classical time-series models.
The fourth paradigm tackles the time irregularity by organizing multiple types of irregular observations into a long sequence of "(time, type, value)" tuples and modeling their interactions via self-attention mechanisms, such as \textbf{SeFT}~\cite{Horn2020SeFunc}, or via graph neural networks, such as \textbf{RainDrop}~\cite{zhang2021raindrop}.
Last, the fifth paradigm refers to recent domain-specific methods for electronic health records (EHRs).
Typical examples include \textbf{AdaCare}~\cite{MaGWZWRTGM20}, a multi-scale model but operating on regularly aggregated time series, and \textbf{STraTS}~\cite{tipirneni2022strats}, a similar variant of \textbf{SeFT} introducing trainable time encodings and auxiliary learning objectives.
We include these two EHR-specific methods in the experiments on large-scale datasets built from \emph{MIMIC-III}.

\subsubsection{\textbf{Implementation Details}}

We employ slightly different hyperparameter configurations across the datasets in our experiments. For a comprehensive reference, we have summarized the hyperparameter settings used for each dataset in Appendix~\ref{app:exp:hyper_param}. Our model is implemented using PyTorch 1.9.0, and both training and inference are performed on CUDA 11.3. We utilize the Adam optimizer with an initial learning rate of $10^{-3}$. The models are trained for a maximum of 50 epochs, and early stopping is applied if there is no improvement on the validation set for 5 consecutive training epochs. All experiments are conducted on NVIDIA Tesla V100 GPUs.
To ensure fair comparisons with other baselines, we explore the hyperparameter search space around their best-reported configurations.

To accommodate the variability in input sequence lengths across datasets, we introduce a normalized length $\tilde{L} = L / L_{data}$, where $L_{data}$ is the median length of training instances. 
When $\tilde{L}$ is less than 1, the warping layer predominantly downsamples the time series, while a value greater than 1 signifies upsampling.
For example, in the case of the \emph{Human Activity} dataset with a median length of $50$, if $\tilde{L}^{(n)}=0.2$, it means that the new sequence length in the $n$-th layer is $10$ (i.e., $10 / 50 = 0.2$). The median lengths for each clinical dataset can be found in Table~\ref{tab:task}.
We utilized specific scales for different datasets in our main experimental results and ablation tests. For the \emph{PhysioNet} dataset, we use three scales with $\tilde{L}^{(1)}=0.2$ and $\tilde{L}^{(2)}=1$. For the \emph{Human Activity} dataset, we use three scales with $\tilde{L}^{(1)}=1.2$ and $\tilde{L}^{(2)}=1$. Lastly, for the \emph{MIMIC-III} based datasets, we utilize two scales with $\tilde{L}^{(1)}=1$.
Given the importance of $\tilde{L}^{(n)}$ as a predefined hyperparameter in our model, we provide a sensitivity analysis in Section~\ref{sec:sensitive} and Appendix~\ref{app:exp:multi_scale} to assess its impact and determine the optimal combinations.


\begin{table}[t]
\centering
\small
\caption{Evaluation results (mean $\pm$ std \%) on \emph{PhysioNet} and \emph{Human Activity}.}
\begin{tabular}{l|cccc}
\toprule
\multirow{2}{*}{\textbf{Model}} & \multicolumn{2}{c}{\textbf{PhysioNet}} & \multicolumn{2}{c}{\textbf{Human Activity}} \\
 & AUROC & AUPRC & Accuracy &  AUPRC \\
\midrule
 RNN-Mean      & 55.5 ± 8.7 & 20.3 ± 7.5	& 74.9 ± 1.4 & 65.5 ± 2.4	 \\ 
 RNN-Forward    & 84.2 ± 0.7 & 52.6 ± 0.7	& 76.7 ± 0.6 & 68.7 ± 0.6	 \\ 
 RNN-$\Delta_t$ & 65.7 ± 3.2 & 25.2 ± 1.5	& 74.6 ± 0.6 & 65.3 ± 2.0	  \\ 
 RNN-Decay & 49.9 ± 0.9 & 15.7 ± 0.4	& 75.9 ± 1.3 & 64.4 ± 1.5	 \\ 
 GRU-D  & 84.7 ± 0.3 & 52.6 ± 0.3	& 75.0 ± 1.0  & 64.3 ± 1.7  \\ 
 Phased-LSTM  & 78.1 ± 0.5 & 40.7 ± 3.6    & 73.5 ± 1.7 & 60.8 ± 1.0   \\ 
 \midrule
 ODE-RNN  & 83.7 ± 0.6 & 53.3 ± 1.1	& 76.0 ± 1.2 & 64.9 ± 2.6  \\ 
 L-ODE-ODE  & 81.2 ± 1.3 & 46.6 ± 3.6	& 76.7 ± 1.4 & 77.1 ± 1.4  \\ 
 \midrule
 SeFT  & 70.8 ± 0.3 & 29.9 ± 0.9  &    75.8 ± 3.3 & 66.1 ± 3.3 \\
 RainDrop & 69.2 ± 4.2 & 28.4 ± 3.6 & 70.2 ± 1.6 & 62.3 ± 2.6 \\
 \midrule
 IP-Nets & 75.6 ± 1.1 & 37.5 ± 2.3	& 73.5 ± 1.7 & 64.7 ± 2.4 \\ 
 mTAND & 86.0 ± 0.4 & 54.6 ± 0.9    & 81.3 ± 0.3 & 73.5 ± 0.8   \\ 
 \midrule
Warpformer      & \textbf{86.6 ± 0.6} &  \textbf{56.7 ± 0.7} & \textbf{84.9 ± 0.7} & \textbf{81.1 ± 0.9}    \\ 
\bottomrule
\end{tabular}
\label{tab:main_phy_act}
\vspace{-0.1in}
\end{table}

\begin{table*}[t]
\centering
\small
\caption{AUROC and AUPRC (mean $\pm$ std \%) of different methods on five datasets built from \emph{MIMIC-III}.}
\begin{tabular}{lcccccccccc}
\toprule
\multirow{2}{*}{\textbf{Model}} &\multicolumn{2}{c}{\textbf{MOR}} & \multicolumn{2}{c}{\textbf{DEC}} & \multicolumn{2}{c}{\textbf{LOS}} & \multicolumn{2}{c}{\textbf{WBM}} & \multicolumn{2}{c}{\textbf{CIP}} \\
 & AUROC & AUPRC & AUROC & AUPRC & AUROC & AUPRC & AUROC & AUPRC & AUROC & AUPRC\\
\midrule
 RNN-Mean       & 89.5 ± 0.3 & 62.9 ± 0.7 & 98.4 ± 0.2 & 83.6 ± 0.8	& 76.7 ± 0.1 & 31.2 ± 0.1	& 79.4 ± 0.3 & 25.8 ± 0.6	& 86.9 ± 0.3 & 45.6 ± 0.2 \\
 RNN-Forward    & 88.5 ± 0.3 & 60.4 ± 1.2 & 97.1 ± 0.6 & 75.6 ± 2.2	& 76.4 ± 0.2 & 30.8 ± 0.1	& 74.6 ± 4.9 & 22.4 ± 3.0	& 86.4 ± 0.4 & 45.2 ± 0.3 \\
 RNN-$\Delta_t$ & 87.4 ± 1.0 & 55.8 ± 2.4 & 97.3 ± 0.5 & 74.5 ± 0.8	& 76.2 ± 0.2 & 30.6 ± 0.2	& 72.5 ± 4.2 & 20.4 ± 2.1	& 50.6 ± 0.6 & 25.2 ± 0.2 \\
 RNN-Decay  & 89.1 ± 0.4 & 62.6 ± 1.0 & 98.4 ± 0.3 & 84.2 ± 2.4	& 77.0 ± 0.1 & 31.6 ± 0.1	& 77.8 ± 2.2 & 24.5 ± 1.9  & 76.2 ± 9.6 & 39.8 ± 4.6 \\
 GRU-D          & 87.6 ± 0.5 & 59.9 ± 0.6 & 97.6 ± 1.0 & 77.9 ± 5.6	& 76.5 ± 0.1 & 30.9 ± 0.3	& 74.4 ± 6.9 & 22.7 ± 4.9	& 84.6 ± 1.8 & 43.3 ± 1.7 \\
 Phased-LSTM    & 86.7 ± 0.3 & 53.7 ± 0.6	& 97.5 ± 0.1 & 78.2 ± 0.4	& 75.6 ± 0.2 & 30.2 ± 0.1	& 77.4 ± 0.3 & 24.3 ± 0.4	& 84.8 ± 0.2 & 44.0 ± 0.1 \\
\midrule
 SeFT      & 88.0 ± 0.4 & 59.3 ± 1.4	& 98.5 ± 0.2 & 86.7 ± 0.7	& 76.1 ± 0.1 & 31.0 ± 0.1	& 83.8 ± 0.4 & 33.1 ± 0.3	& 86.6 ± 0.2 & 45.0 ± 0.1 \\
 RainDrop     & 87.6 ± 0.1 & 59.1 ± 0.4	& 98.0 ± 0.2 & 82.7 ± 0.5	& 76.0 ± 0.2 & 30.6 ± 0.4	& 80.4 ± 0.4 & 27.7 ± 0.6	& 86.9 ± 0.4 & 44.8 ± 0.2 \\
 \midrule
 IP-Nets     & 88.9 ± 0.3 & 61.9 ± 0.9	& 98.3 ± 0.1 & 85.2 ± 0.6	& 74.3 ± 3.6 & 28.8 ± 3.7	& 81.6 ± 0.1 & 28.1 ± 0.6	& 85.2 ± 0.6 & 44.0 ± 0.6 \\
 mTAND      & 89.0 ± 0.2 & 61.8 ± 0.7	& 97.2 ± 0.3 & 74.5 ± 3.2	& 73.8 ± 0.4 & 28.3 ± 0.4	& 66.1 ± 0.2 & 16.7 ± 0.1	& 84.2 ± 0.3 & 42.1 ± 0.2 \\
\midrule
 AdaCare    & 76.0 ± 0.5 & 47.7 ± 0.7	& 94.9 ± 2.0 & 68.5 ± 9.0	& 64.9 ± 0.6 & 22.1 ± 1.2	& 51.2 ± 0.2 & 12.9 ± 0.0	& 64.4 ± 2.4 & 36.2 ± 1.9 \\
 STraTS    & 89.3 ± 0.1 & 61.3 ± 0.3	& 98.6 ± 0.1 & 84.0 ± 1.6	& 76.6 ± 0.2 & 31.4 ± 0.3	& 79.1 ± 2.1 & 26.2 ± 2.4	& 87.8 ± 0.2 & 45.6 ± 0.2 \\
 \midrule
Warpformer      & \textbf{90.3 ± 0.1} & \textbf{64.6 ± 0.4}	& \textbf{99.0 ± 0.1} & \textbf{90.0 ± 0.4}	& \textbf{77.7 ± 0.2} & \textbf{32.5 ± 0.2}	& \textbf{85.5 ± 0.1} & \textbf{35.5 ± 0.3}	& \textbf{88.0 ± 0.2} & \textbf{46.4 ± 0.2} \\
\bottomrule
\end{tabular}
\label{tab:auprc_mimic3}
\vspace{-0.1in}
\end{table*}

\subsection{Experimental Results}
\label{sec:exp_res}

\subsubsection{\textbf{Main Results}}
We include the overall comparison results on \emph{PhysioNet} and \emph{Human Activity} in Table~\ref{tab:main_phy_act}.
Besides, we present AUROC and AUPRC results on \emph{MIMIC-III}-based datasets in Table~\ref{tab:auprc_mimic3}.
In general, we observe that different baselines present apparent performance variations across different datasets and evaluation metrics.
Interestingly, certain advanced methods, such as mTAND, demonstrate excellent performance on the \emph{PhysioNet} and \emph{Human Activity} datasets. However, despite extensive hyper-parameter tuning, these methods do not exhibit comparable performance on the \emph{MIMIC-III}-based datasets. On the other hand, basic extensions of recurrent neural networks, such as RNN-Mean and RNN-Decay, perform poorly on \emph{PhysioNet} but surprisingly achieve competitive and robust results on \emph{MOR}, \emph{DEC}, and \emph{LOS} tasks. 
We conjecture the underlying reason is related to the matching degree between the effective patterns of a specific dataset and the unique assumptions held by a specific model.
Nevertheless, with the capability of encoding various crucial information in irregular time series and unifying both fine-grained and coarse-grained information, Warpformer achieves remarkable performance improvements over the most competitive baselines in all setups.

Additionally, we observed that the improvements achieved by Warpformer vary across tasks, with specific tasks, e.g., WBM, benefiting more than others, e.g., MOR. One possible explanation for this variation is the inherent difficulty of tasks. The WBM task involves predicting a substantial number of biomarkers to be measured in the upcoming hour, necessitating the model to capture inter-variable discrepancies. In contrast, the MOR task only involves binary classification, which is relatively straightforward and demands less attention to inter-variable discrepancies. The increased complexity and inter-variable challenges in the WBM task likely contribute to the significant gains achieved by the Warpformer model compared to the MOR task.
Another factor that can impact performance is the length of observation windows for each task. Shorter observation windows may limit the model's ability to capture complex temporal patterns and obscure intra- and inter-variable discrepancies. For example, the WBM task requires a 48-hour look-back window, necessitating the model to capture intricate temporal patterns. In contrast, the CIP task only requires a 6-hour look-back window, which could explain its minor improvement compared to the tasks with longer observation windows.

\begin{table}[ht]
\centering
\small
\caption{Ablation tests of Warpformer on \emph{PhysioNet} and \emph{Human Activity}.}
\begin{tabular}{l|cccc}
\toprule
\multirow{2}{*}{\textbf{Warpformer}} & \multicolumn{2}{c}{\textbf{PhysioNet}} & \multicolumn{2}{c}{\textbf{Human Activity}}  \\
 & AUROC & AUPRC & Accuracy & AUPRC   \\
\midrule
Full & \textbf{86.6 ± 0.6} &  \textbf{56.7 ± 0.7} & \textbf{84.9 ± 0.7} & \textbf{81.1 ± 0.9} \\
\midrule
No Up-sampling & 85.4 ± 1.2 & 54.5 ± 1.8 & 84.4 ± 0.8 & 79.0 ± 0.7 \\
Identical Map. & 84.8 ± 0.4 & 52.0 ± 2.1 & 84.8 ± 1.2 & 80.0 ± 1.4  \\
Hourly Agg. & 84.2 ± 0.5 & 51.6 ± 2.4 & N/A & N/A  \\
\midrule
$-f^{\textrm{abs}} (\cdot)$ & 83.2 ± 1.3 & 50.9 ± 1.6 & 80.6 ± 1.9 & 74.0 ± 2.3 \\
$-f^{\textrm{rel}} (\cdot)$ & 86.3 ± 1.1 & 54.5 ± 2.7 & 84.8 ± 0.7 & 80.5 ± 1.0 \\
$-f^{\textrm{type}} (\cdot)$ & 86.3 ± 0.9 & 55.7 ± 2.3 & 84.3 ± 0.4 & 79.3 ± 0.9 \\
\midrule
Input Pooling & 47.3 ± 0.1 & 14.8 ± 0.0 & 32.4 ± 0.0 & 15.7 ± 0.1 \\
\bottomrule
\end{tabular}
\label{tab:abl_warp_phy_act}
\vspace{-0.1in}
\end{table}

\subsubsection{\textbf{Ablation Tests}}
We conducted ablation tests on Warpformer to illustrate the significance of critical designs. 
Table~\ref{tab:abl_warp_phy_act} includes the results of three warping-related variants, 1) disabling the up-sampling functionality (No Up-sampling); 2) using the identical mapping to substitute the warping-based mapping (Identical Mapping); 3) using hourly aggregation to substitute the warping module (Hourly Aggregation).
The results demonstrate that the warping module plays an important role compared to identical mapping. Notably, the ability to upsample significantly impacts the performance, as the absence of upsampling functionality can even lead to inferior results compared to identical mapping. These findings underscore the necessity of capturing finer-grained details of sparsely sampled signals during multi-scale modeling.
Furthermore, the experimental results on the \emph{PhysioNet} dataset reveal that inappropriate schemes for obtaining coarse-grained representations, such as aggregating all data within a time slot, can yield worse results than a model without a multi-scale setting. These observations validate the effectiveness of the adaptive unification strategy employed.

To further explore the importance of different components in the input encoding process, we conducted additional ablation tests, and the results are also presented in Table~\ref{tab:abl_warp_phy_act}. The notation $-f^{\textrm{abs}}(\cdot)$, $-f^{\textrm{rel}}(\cdot)$, and $-f^{\textrm{type}}(\cdot)$ represent the removal of absolute-time embeddings, relative-time embeddings, and type embeddings, respectively. The term "Input pooling" refers to the application of a simple average pooling operation that collapses the temporal and variate dimensions.
The results demonstrate that the model experiences a significant performance drop when any encoding components are missing. Particularly, the absence of absolute-time encoding has the most pronounced impact on the model's performance. This finding is intuitive as $f^{\textrm{abs}}(\cdot)$ captures the absolute position of each sample within a continuous time period, which is crucial for effectively modeling irregularly sampled time series.
These ablation tests provide strong evidence supporting the importance of each encoding component in the input representation of Warpformer. By considering these components together, Warpformer is able to capture the intricate temporal relationships and characteristics of irregularly sampled time series data.


\begin{figure*}[ht]
  \centering
    \subfloat[\emph{PhysioNet} (AUROC, 2 scales)]{
    \label{fig:physio1D_auroc}
    \includegraphics[width=0.25\linewidth]{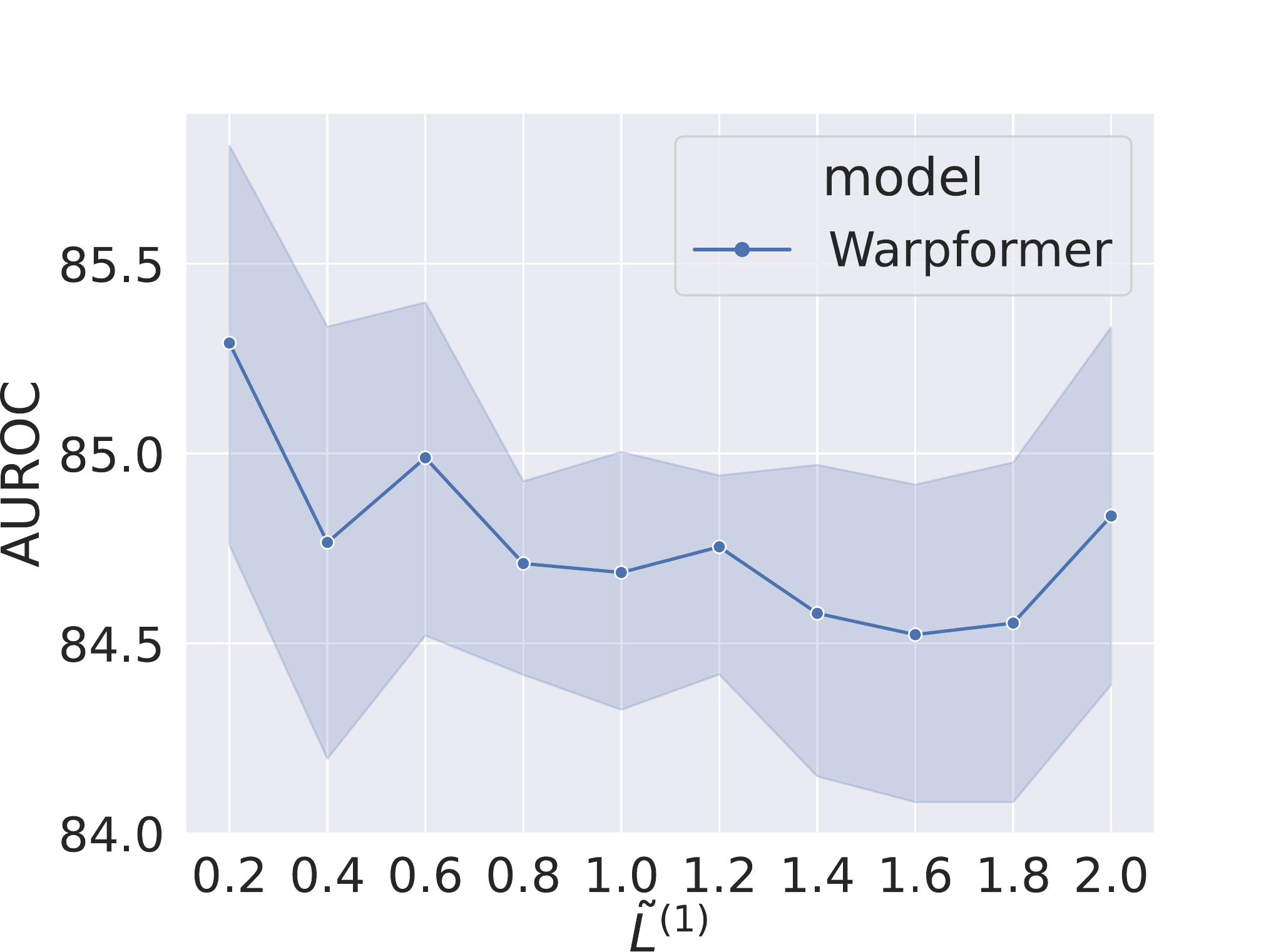}}
    \subfloat[\emph{Human Activity} (ACC, 2 scales)]{
    \label{fig:active1d_acc}
    \includegraphics[width=0.25\linewidth]{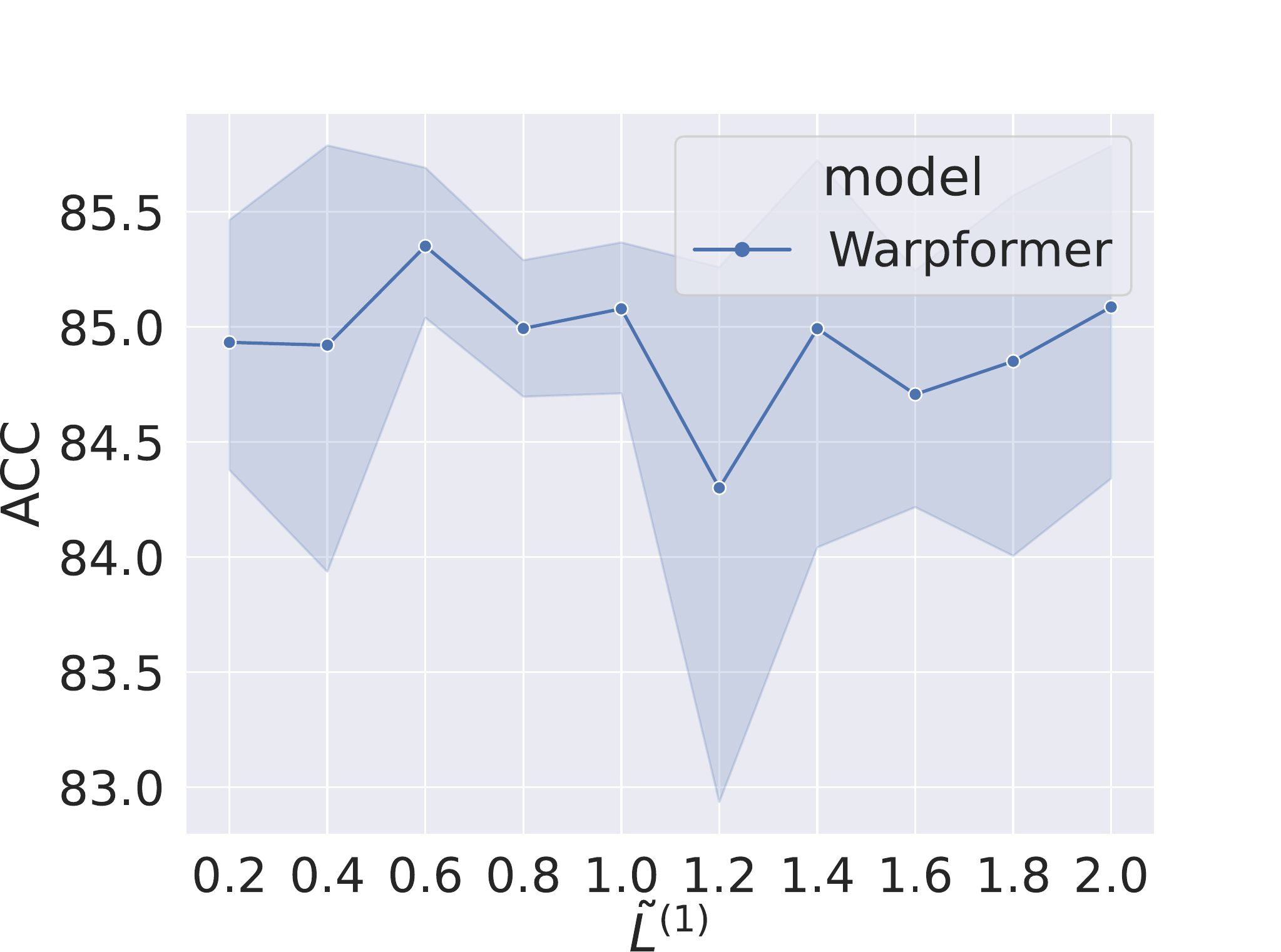}}
    \subfloat[\emph{PhysioNet} (AUROC, 3 scales)]{
    \label{fig:physio2D_auroc}
    \includegraphics[width=0.24\linewidth]{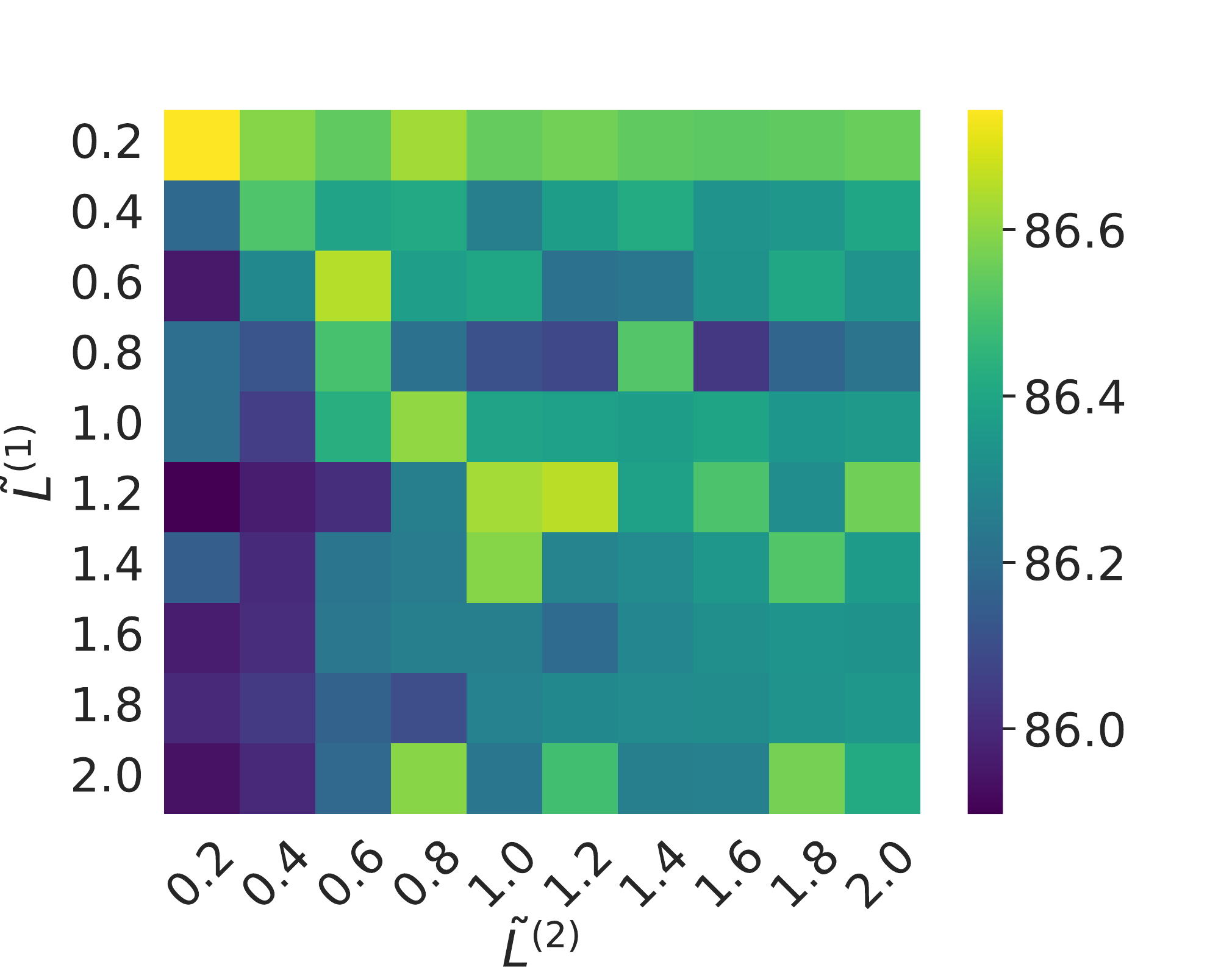}}
    \subfloat[\emph{Human Activity} (ACC, 3 scales)]{
    \label{fig:active2d_acc}
    \includegraphics[width=0.24\linewidth]{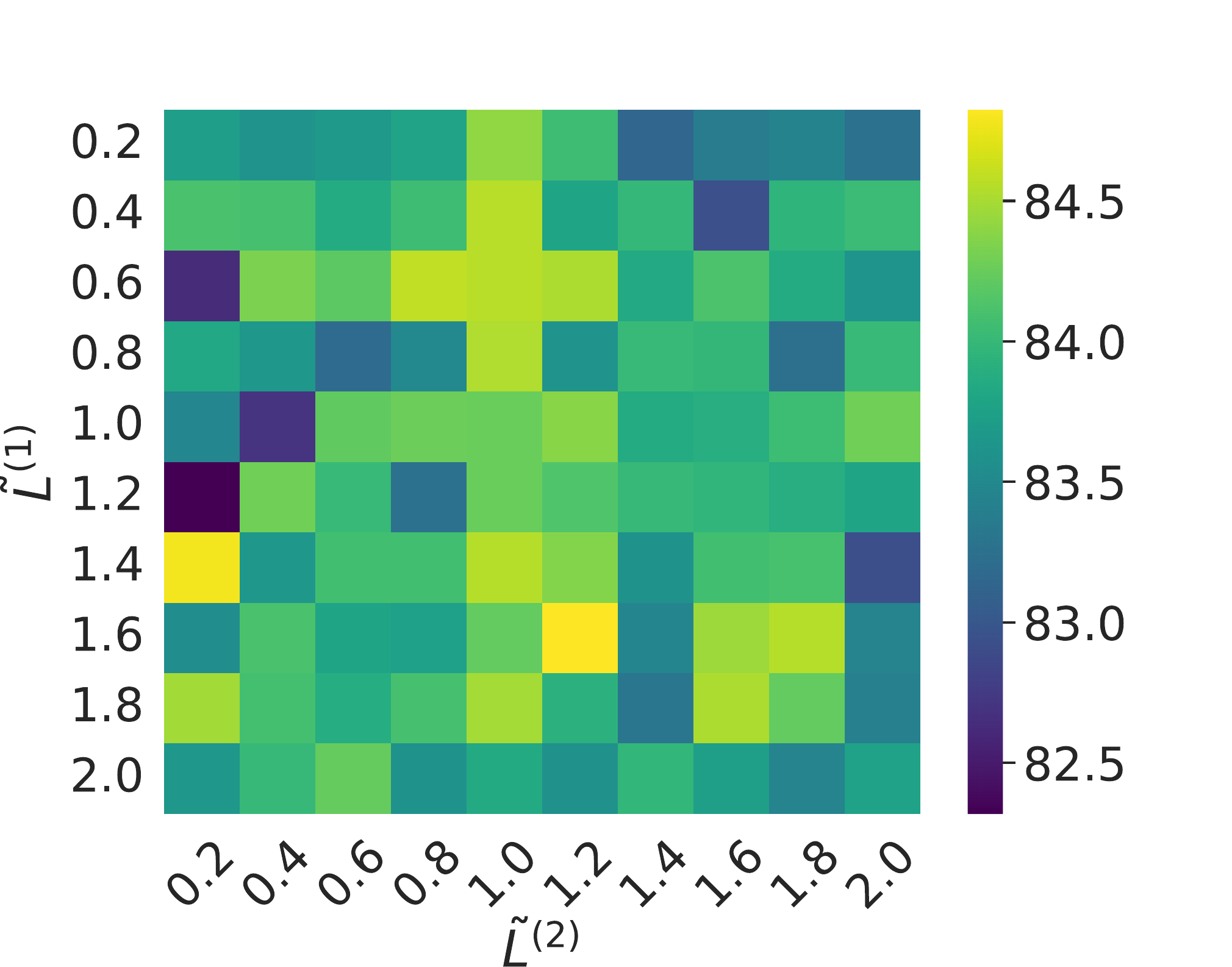}}
  \caption{\label{fig:sensitive} The multi-scale effects of Warpformer on \emph{PhysioNet} and \emph{Human Activity}.
  }
\end{figure*}

\begin{figure*}[ht]
  \centering
  \includegraphics[width=\textwidth]{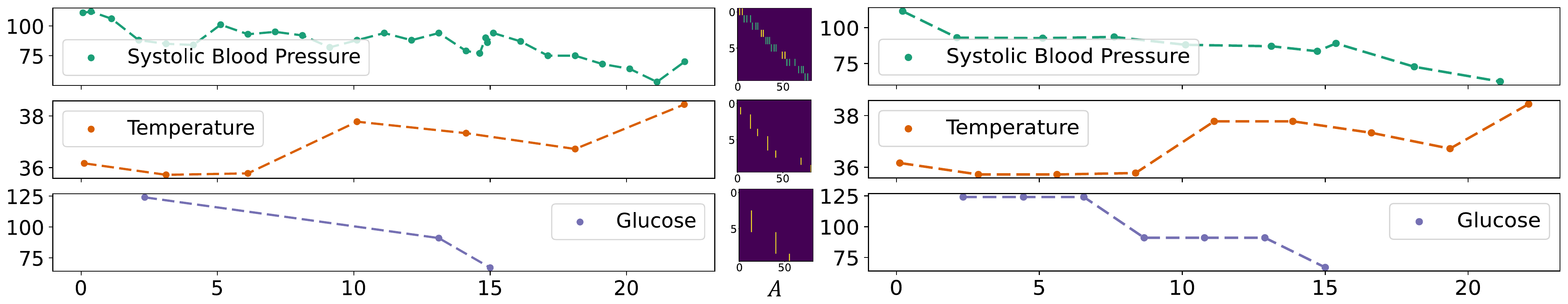}
  \caption{Visualization of the input time series (left) and its adaptive unification (right), as well as the learned corresponding alignment matrix $A$ (middle). The length of the y-axis of $A$ is $\tilde{L}^{(n)}$, i.e., the size of the produced unification, and the x-axis is $\tilde{L}^{(n-1)}$, i.e., the length of the input time series.}
  \label{fig:viz}
\end{figure*}

\subsubsection{\textbf{Multi-scale Effects}}
\label{sec:sensitive}
We study the multi-scale effects of Warpformer with different setups of the number of Warpformer layers $N$, i.e., the number of scales, and the hyper-parameters $\{\tilde{L}^{(n)}\}_{n=1}^N$, i.e., the unification granularity.
Figure~\ref{fig:sensitive} includes the results on \emph{PhysioNet} and \emph{Human Activity}, where we show the performance as a function of $\tilde{L}^{(n)}$.
Note that in the case of the \emph{Human Activity} dataset, hidden states must be generated for each time step since it involves per-step prediction. Consequently, the 2 scales mentioned in Figure~\ref{fig:active1d_acc} actually correspond to 3 Warpformer layers ($N=2$), where $\tilde{L}^{(2)}=1$. Similarly, the 3 scales mentioned in Figure~\ref{fig:active2d_acc} correspond to 4 Warpformer layers ($N=3$).

These findings suggest that the optimal number of layers applied varies across different datasets. The optimal number is often not very large, indicating that using 2 or 3 scales in Warpformer for encoding is sufficient to generate highly effective representations for downstream tasks. We observed that increasing the number of layers, which involves repeating up- or down-sampling operations, did not yield substantial additional benefits. In fact, it was noticed that excessive stacking of scales could potentially weaken the patterns learned in the previous layers. This observation emphasizes the importance of striking the right balance in the number of scales used, avoiding unnecessary complexity that could impede the model's ability to capture and leverage meaningful information. 

Moreover, the results highlight the significance of selecting an appropriate unification granularity in Warpformer. Interestingly, different datasets showed distinct preferences for down-sampling and up-sampling configurations. To be specific, extreme down-sampling had the most substantial performance improvement on the \emph{PhysioNet} dataset, whereas the \emph{Human Activity} dataset preferred up-sampling configurations.
We speculate that the disparity in performance can be attributed to the nature of the prediction tasks in each dataset. The \emph{PhysioNet} dataset involves a single prediction target for the entire series, prioritizing the capture of long-term trends and overall patterns. Thus, a more coarse-grained representation proves beneficial in this context. On the other hand, the \emph{Human Activity} dataset requires accurate predictions at each time point, necessitating fine-grained information to discern subtle variations and dependencies within the temporal sequences. Therefore, the up-sampling configurations are particularly advantageous for this dataset.
These findings shed light on the importance of adapting the unification granularity to the specific characteristics of the dataset and prediction task. By tailoring the level of down-sampling or up-sampling, Warpformer can effectively capture the relevant temporal patterns and optimize its performance accordingly.

\subsection{Case Study}
\label{sec:case_study}

To gain deeper insights into how the warping mechanism enhances predictions in downstream tasks, we conducted a comprehensive case study focusing on instances that exhibited substantial improvements after the warping process. The visual analysis, as presented in Figure~\ref{fig:viz}, includes three representative physiological signals from the MIMIC-III-based datasets, showcasing the original input signals, the corresponding learned alignment matrix $A$, and the output series after adaptive unification.

The visualizations demonstrate that the warped signals faithfully capture the trends present in the original signals, and the warping module exhibits the ability to compute variate-specific alignment schemas. For densely sampled signals, such as systolic blood pressure, the warping module downsamples the original series while preserving irregular patterns and important local fluctuations. This is achieved by retaining the intra-variate irregularity during the unification process, which cannot be achieved through traditional hourly aggregation. Such capability is advantageous in multi-scale modeling as it prevents the loss of irregular time patterns at coarse-grained levels and retains valuable details.
On the other hand, for sparsely sampled signals, e.g., body temperature and glucose, the warping module generates additional interpolated sample points, resulting in a more fine-grained representation of the coarse-grained signal. These interpolated sample points provide more detailed features, facilitating the model's ability to capture subtle variations in the temporal sequences.

The case study clearly demonstrates that the warping mechanism successfully preserves intra-variate irregularities and alleviates inter-variate discrepancies. This underscores the crucial role of the adaptive unification process in Warpformer, which significantly contributes to the model's performance and ability to capture the underlying dynamics of the irregular clinical time series.

%% file: conclusion.tex
\section{Conclusion and Future Work}
\label{sec:conclu}

In conclusion, our paper has highlighted the importance of addressing both intra-series irregularity and inter-series discrepancy when dealing with irregularly sampled multivariate time series, particularly in clinical scenarios.
We have presented the first multi-scale approach, Warpformer, which has achieved remarkable performance across various datasets.
However, it is important to acknowledge that our work has several limitations at this stage.
Firstly, the reliance on pre-specified hyper-parameters, such as the number of layers and per-layer unification granularity, poses a challenge in terms of efficiency in hyper-parameter tuning.
Secondly, the maintenance of both temporal and variate dimensions in internal feature representations introduces additional space complexity.
Finally, Warpformer's dependence on gradients from downstream tasks limits its use to supervised settings, which may affect its generalization performance in real-world scenarios with limited labeled data.
Future work will explore ways to address these limitations and improve the efficiency and generalization of our approach.

%% file: supplementary.tex
\section{MIMIC-III-based datasets}
\label{app:data:mimic3}

\begin{figure}[htb]
  \centering
    \subfloat[LOS.]{
    \label{fig:los_label}
    \includegraphics[width=0.48\linewidth]{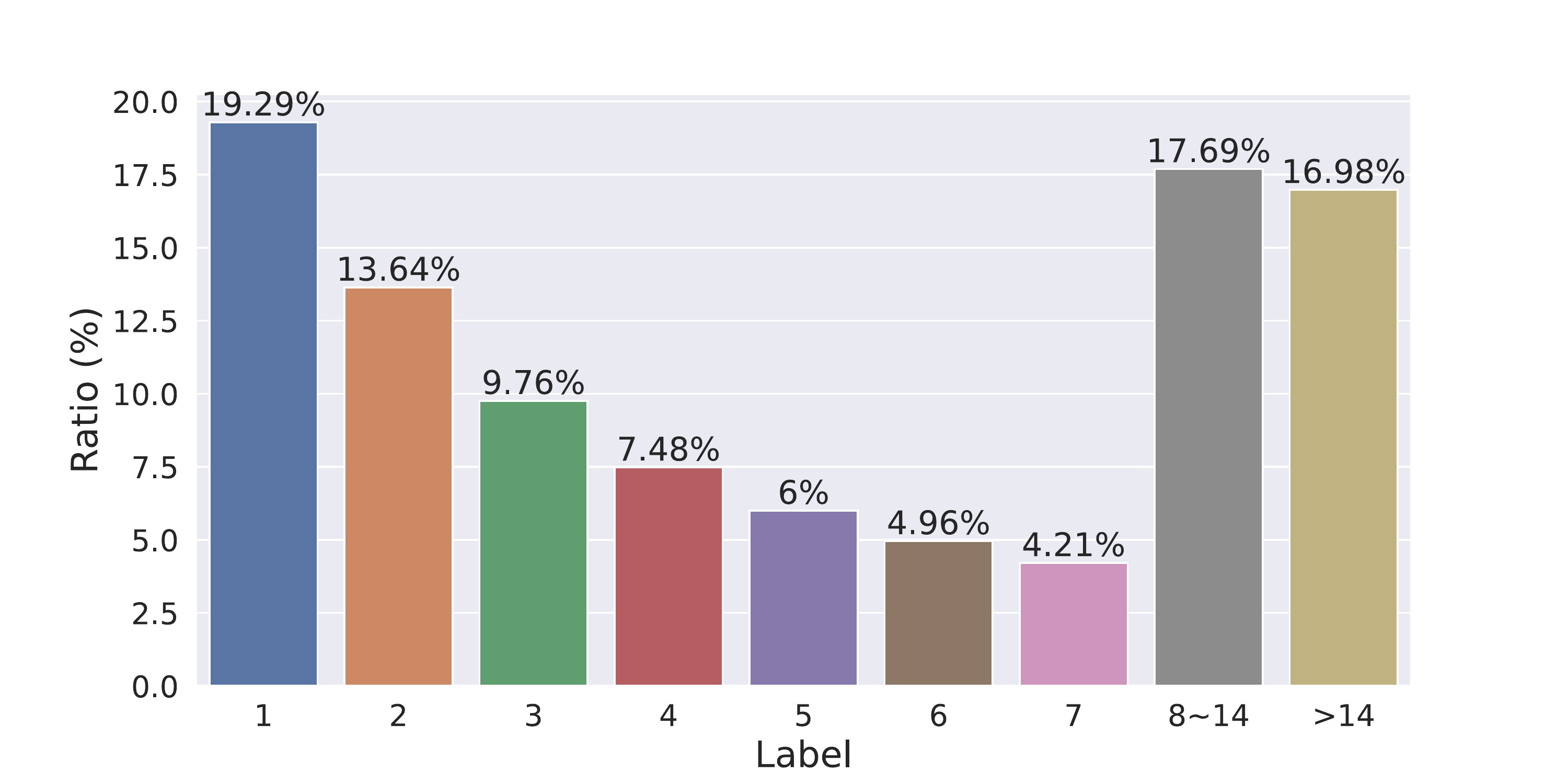}}
    \subfloat[CIP.]{
    \label{fig:cip_label}
    \includegraphics[width=0.48\linewidth]{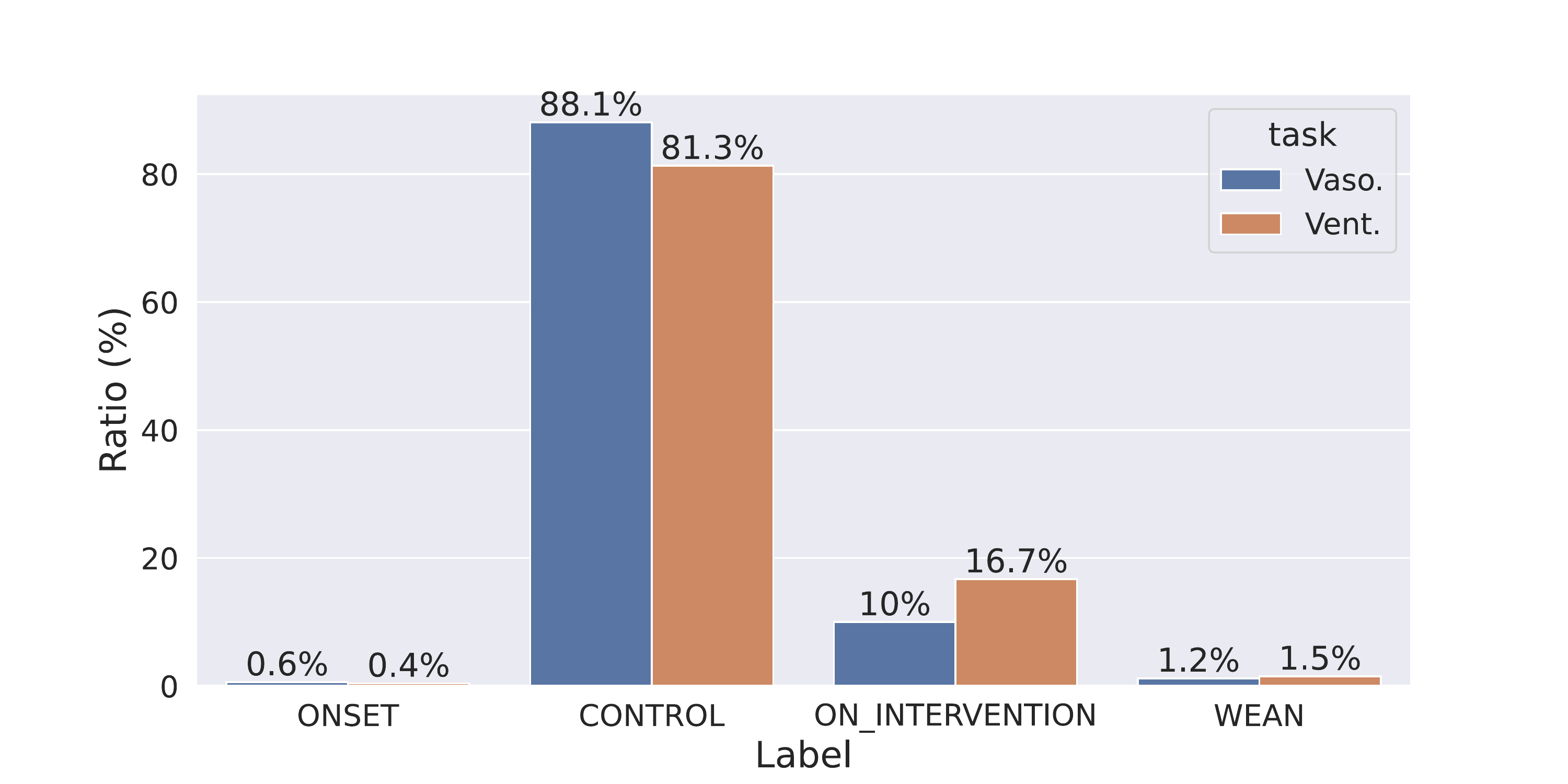}}
    
    \subfloat[WBM.]{
    \label{fig:wbm_label}
    \includegraphics[width=0.98\linewidth]{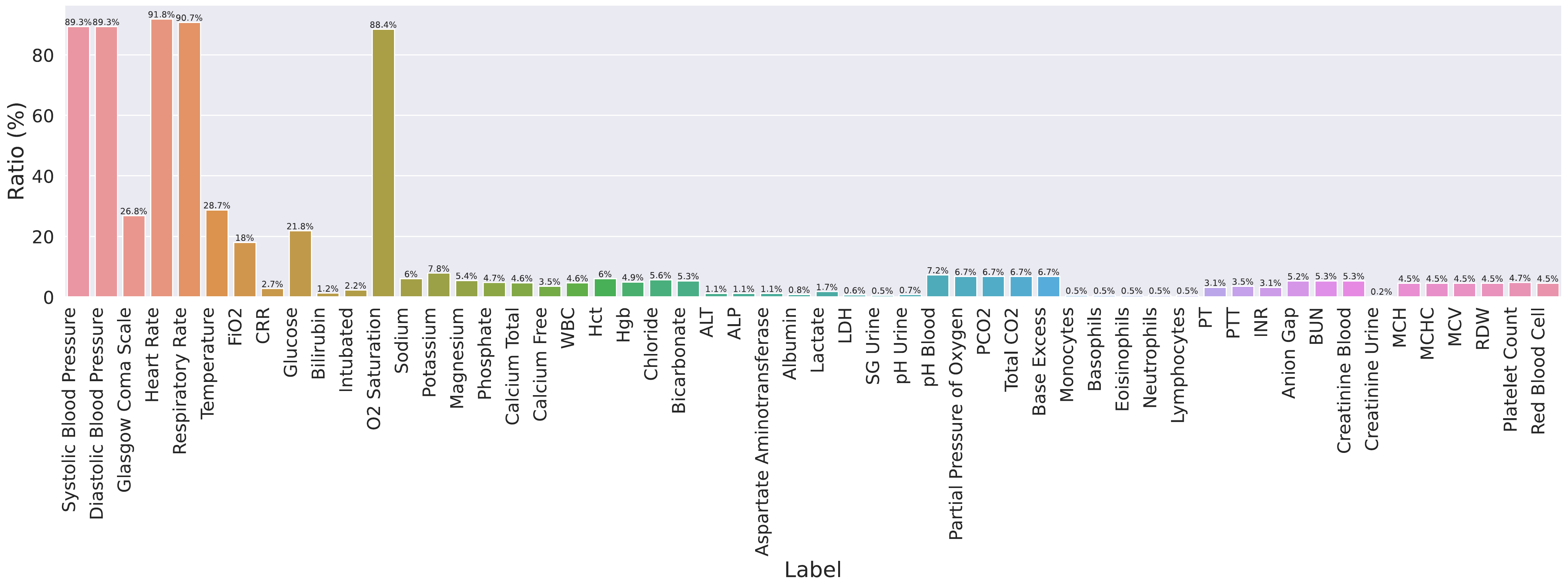}}
  \caption{\label{fig:mimic_label} Label distribution of LOS, CIP, and WBM datasets.}
\end{figure}

\begin{figure*}[htb]
  \centering
  \includegraphics[width=0.9\textwidth]{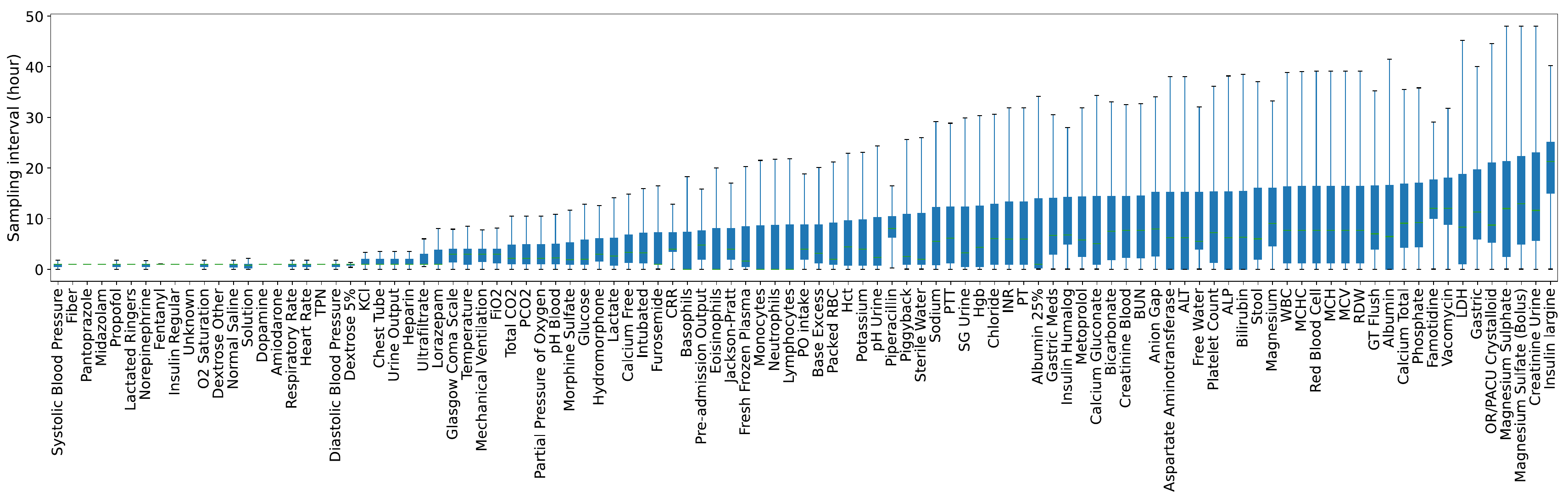}
  \caption{\label{fig:total_boxplot} The distribution of sampling intervals for all the clinical signals in the MOR task.}
\end{figure*}

\emph{MIMIC-III} database~\cite{johnson2016mimic} contains hospital admission records of $53,423$ de-identified patients.
We selected $61$ common biomarker variables and $42$ widely used interventions in intensive care units (ICU)~\cite{tipirneni2022strats}. We also follow the previous studies~\cite{Shukla2019IP-Net,MIMIC-Extract} to ensure the sufficient duration of hospital stay by removing the records with the length of stay less than $48$ hours.
Figure~\ref{fig:total_boxplot} illustrates the distribution of sampling intervals in the in-hospital mortality task for all clinical signals, visually representing the wide range of variate and significant inter-variate discrepancies in the clinical time series. 
To imitate the practical situation, we sort the clinical records according to the time of admission, where the training set includes the earliest 80\% of admission records.
We equally divide the rest into validation and testing sets.
There is no admission overlap in different sets to avoid data leakage issues.
Similar to~\cite{Shukla2019IP-Net,MIMIC-Extract,McDermott2021chil}, we select a diverse set of tasks to cover different clinical scenarios. 
Table~\ref{tab:task} summarizes the statistics of these tasks, which will be briefly introduced one by one in the following.

\textbf{In-hospital Mortality (MOR)}
The goal of the MOR task is to predict whether a patient will decease at the end of this hospital stay. In this task, models can observe only the first $48$ hours, which is advantageous from a practical view since the sooner clinicians identify the risks, the more prompt they can implement interventions.
As performed in existing studies~\cite{Shukla2019IP-Net, shukla2021multitime, MIMIC-Extract}, we define this task as a binary classification problem, where 1 indicates that the patient deceased in this hospital stay, otherwise 0.
The overall in-hospital mortality rate of this benchmark is 11.66\%.

\textbf{Decompensation (DEC)}
The objective of DEC is to determine whether a patient will decease in the next $24$ hours based on the data within a $12$-hour time window.
This task, also referred to as imminent mortality prediction, has been studied widely as a signal for more general physiological decompensation~\cite{harutyunyan2019multitask, McDermott2021chil}.
We formulate this task as a binary classification as well, but sampling data throughout the entire hospital stay in the rolling form. 
In this way, we obtained $311, 161$ samples with a mortality rate of $1.81\%$.

\textbf{Length Of Stay (LOS)}
Compared with other tasks, the LOS prediction provides a more fine-grained view of patients' physiological states, helping clinicians better monitor the progress of the disease.
As did in~\cite{xu2018raim}, we formulate this task as a classification problem with $9$ classes, where class 1-7 corresponds to 1-7 days of stay, respectively, class 8 for more than eight days but less than two weeks of stay, and class 9 for living over two weeks.
As a rolling task, its observation is sampled every $12$ hours throughout the entire stay, where the sliding window size is $24$ hours.
The distribution of labels for this dataset can be found in Figure~\ref{fig:los_label}.

\textbf{Next Timepoint Will Be Measured (WBM)}
The WBM task is a multi-label classification problem with the goal of predicting which biomarker signals will be measured in the next hour~\cite{McDermott2021chil}.
This task is beneficial to assist clinicians in developing the subsequent treatment plan.
To fulfill this task, the model needs to make predictions for $54$ biomarkers, each of which is a binary classification problem. If a signal was measured within the following hour, it is assigned a label of 1, otherwise 0.
The WBM task is conducted over the entire stay of a patient, where the sampling stride is $12$ hours and the observation window size is $48$ hours.
We provide the distributions of each prediction target in Figure~\ref{fig:wbm_label}.

\textbf{Clinical Intervention Prediction (CIP)}
As a critical clinical application, accurately predicting the clinical interventions can largely release the burden of clinicians in practice~\cite{wu2017AMIA, Ghassemi2017CRI}.
Similar to~\cite{suresh2017clinical, MIMIC-Extract}, we also formulate the CIP task as a multi-class classification problem.
This task consists of two subtasks in terms of different intervention types: \emph{mechanical ventilation} and \emph{vasopressor}.
The prediction target for each type should be one of the four possible options: 1) onset, 2) wean, 3) continuing on the intervention, and 4) continuing to stay off the intervention.
For more details on the distribution of prediction targets, please refer to Figure~\ref{fig:cip_label}.
The input of this task is the historical data within the 6-hour lookback window, and the prediction target is the state of the intervention within the next 4-hour lookahead window.
The CIP task is also conducted over the entire stay of a patient, where the sampling stride is $6$ hours.


\begin{figure}[b]
  \centering
    \subfloat[MIMIC-III (AUROC, 2 scales)]{
    \label{fig:mimic2d_auroc}
    \includegraphics[width=0.48\linewidth]{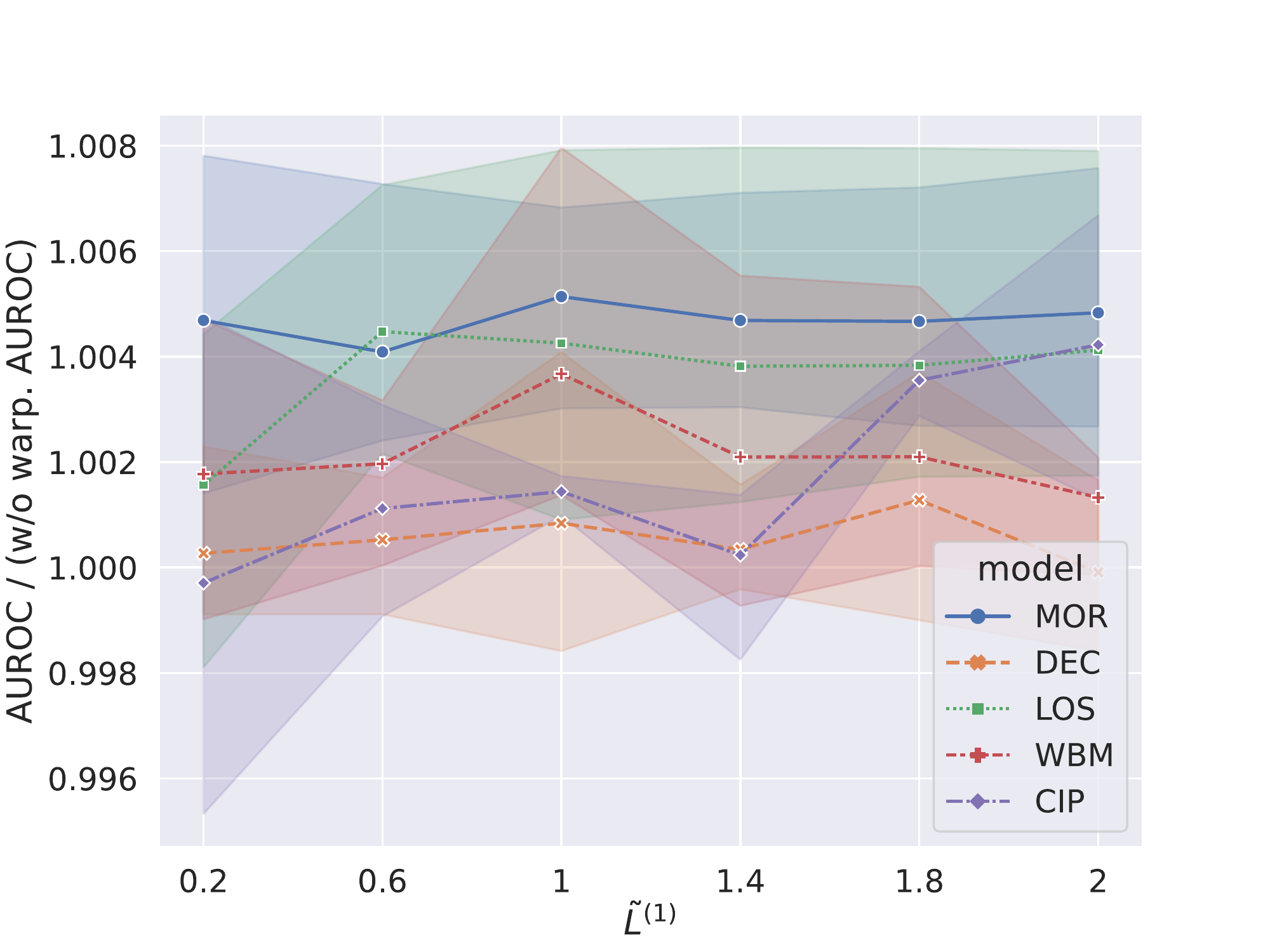}}
    \subfloat[MIMIC-III (AUPRC, 2 scales)]{
    \label{fig:mimic2d_auprc}
    \includegraphics[width=0.48\linewidth]{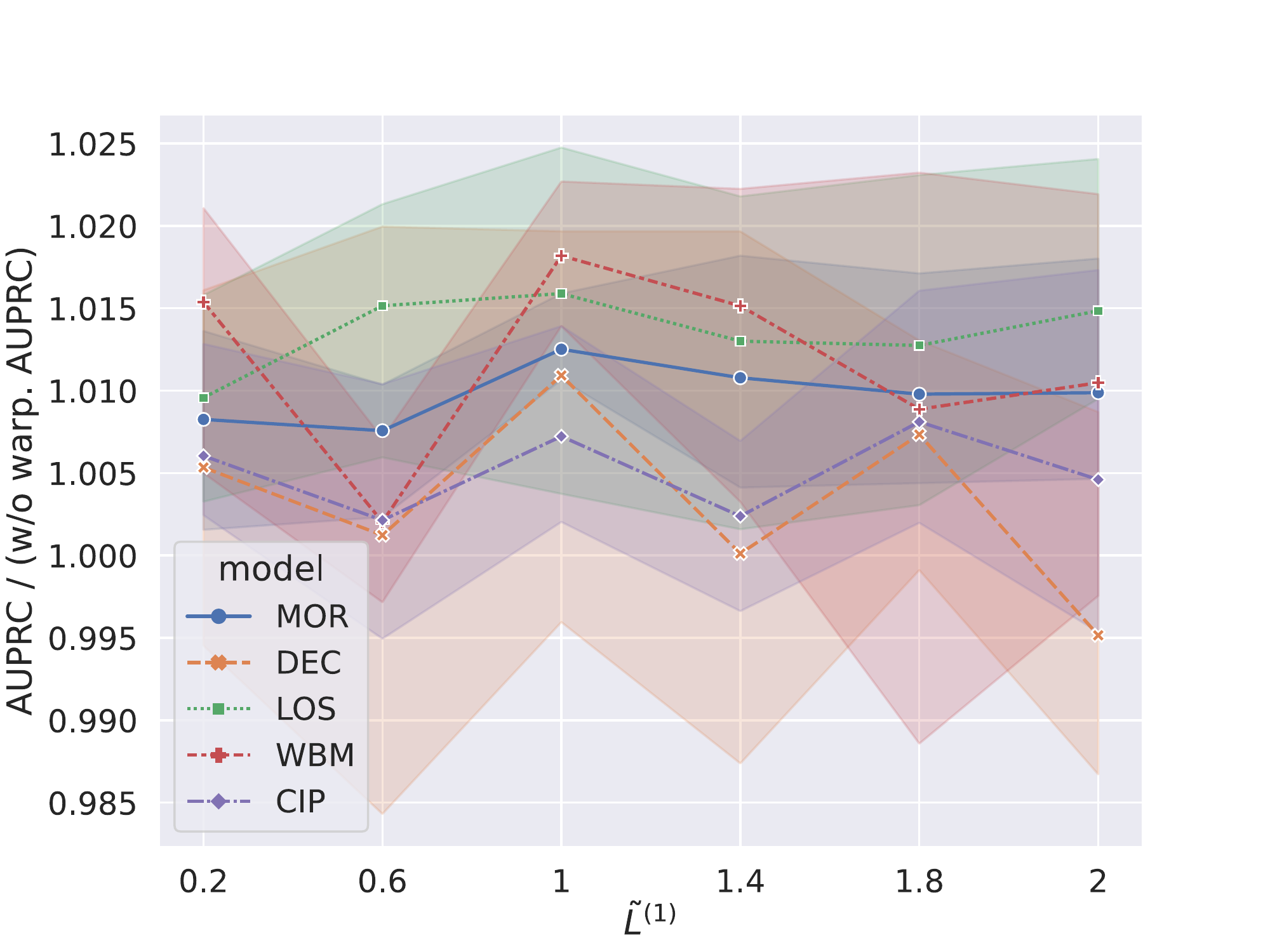}}
  \caption{\label{fig:sensitive1d} Performance of Warpformer with varying $\tilde{L}^{(1)}$ on MIMIC-III-based datasets.}
\end{figure}

\begin{table*}[htb]
\centering
\small
\caption{\label{tab:task} Specifications for five clinical datasets derived from the \emph{MIMIC-III} database (\textit{BC}: binary classification, \textit{ML}: multi-label classification, \textit{MC}: multi-class classification).}
\begin{tabular}{lcrrrcl}
\toprule
\textbf{Task} \textbf{(Abbr.)} & \textbf{Type} & \textbf{\# Train}  & \textbf{\# Val.} & \textbf{\# Test} & \textbf{Median Seq. Len.} & \textbf{Clinical Scenario} \\ \midrule
In-hospital Mortality  (MOR)      & BC  & 39, 449    & 4, 939  & 4, 970   & 63 & Early warning \\
Decompensation   (DEC)    & BC  & 249, 045    & 31, 896  & 30, 220   & 34 & Outcome prediction \\
Length Of Stay   (LOS)       & MC    & 249, 572    & 31, 970  & 30, 283  & 34 & Outcome prediction \\
Next Timepoint Will Be Measured (WBM)   & ML    & 223, 867  & 28, 754 & 27, 038  & 78 & Treatment recommendation \\
Clinical Intervention Prediction (CIP)   & MC    & 223, 913  & 28, 069  & 27, 285 & 10 & Treatment recommendation \\
\bottomrule
\end{tabular}
\end{table*}

\begin{table}[ht]
\centering
\caption{\label{tab:hyperpara} Specifications for hyper-parameters.}
\begin{tabular}{lcccc}
\toprule
\textbf{Dataset} & $D$ & \textbf{Batch size}  & \textbf{\# Head} & \textbf{\# Layers} $J$ \\ 
\midrule
PhysioNet       & 32 & 32 & 1 & 2 \\
Human Activity  & 64 & 64 & 8 & 3 \\
MIMIC-III       & 32 & 32 & 1 & 2 \\
\bottomrule
\end{tabular}
\end{table}

\section{Experiments}
\label{app:exp}

\subsection{Performance Metrics}
\label{app:exp:metrics}

Clinical downstream tasks, e.g., mortality prediction, frequently encounter significant data imbalance issues. Consequently, we employed AUROC and AUPRC as the primary evaluation metrics. Due to the balanced nature of the \emph{Human Activity} dataset and the prevailing usage of Accuracy in previous studies~\cite{Shukla2019IP-Net,shukla2021multitime}, we also adopted Accuracy for this particular dataset. For each dataset, we select the model parameters that yield the highest AUROC value on the validation set and apply them for evaluation on the test set.

\subsection{Further Details on Hyper-parameters}
\label{app:exp:hyper_param}

For each dataset, we tailor the dimensionality of hidden states, batch size, and the number of heads and layers in the doubly attention module. Please refer to Table~\ref{tab:hyperpara} for specific configuration details.
In our implementation, each attention head is set to have a dimension of 8. The score function $f^{\bm{s}}(\cdot)$ is implemented as a two-layer fully connected network with a ReLU activation function. Through thorough testing, we find that using the Sigmoid function to compute the non-negative score ensures better stability in our model.

\subsection{Further analyses on multi-scale hyper-parameters}
\label{app:exp:multi_scale}

\begin{figure}[hb]
  \centering
  \includegraphics[width=0.9\linewidth]{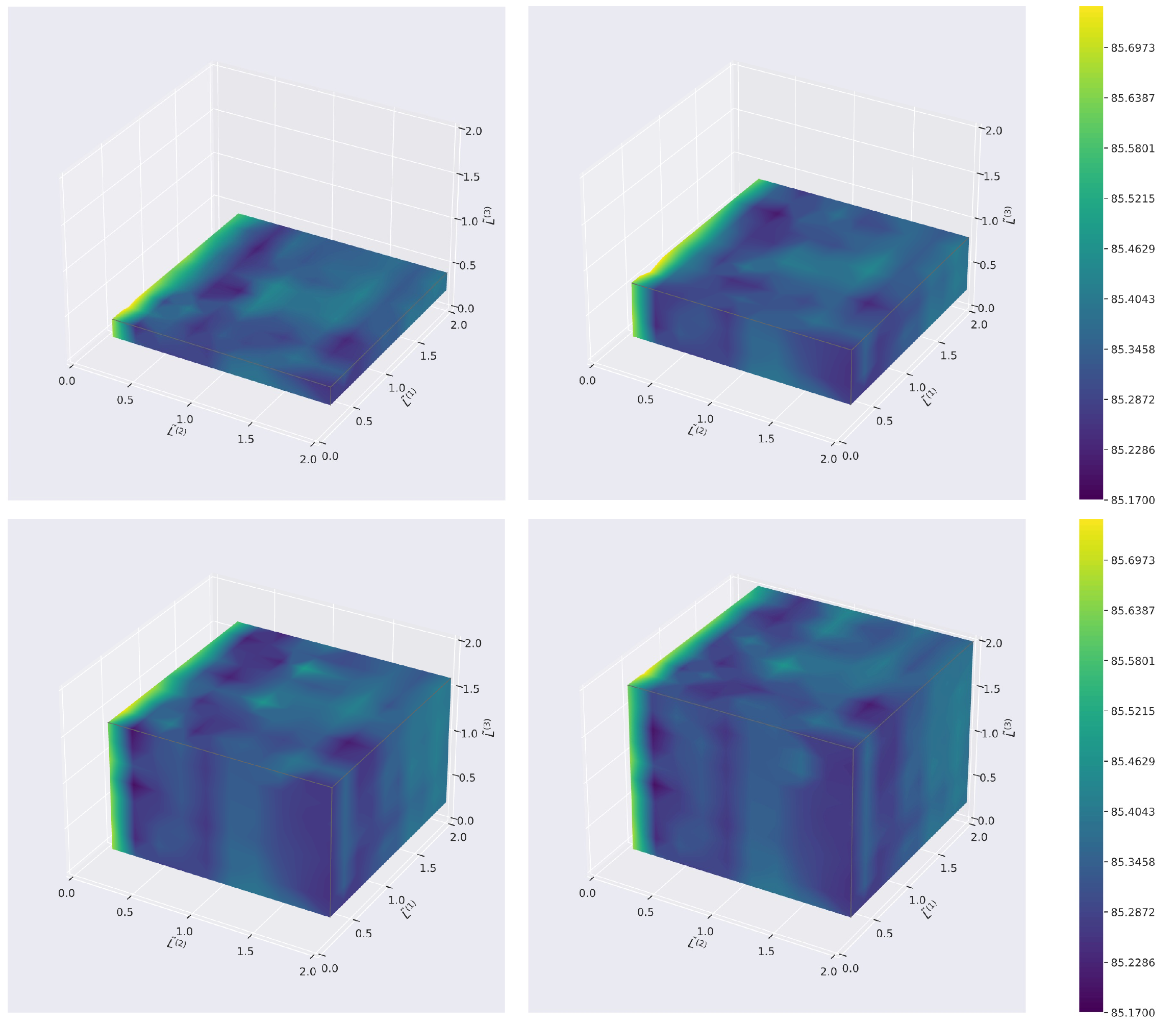}
  \caption{\label{fig:physio3d_auroc} AUROC of the four-layer Warpformer on the PhysioNet dataset with different $\tilde{L}^{(n)}$ settings.}
\end{figure}

To further demonstrate the impact of scale numbers ($N$) and unification granularity ($\tilde{L}^{(n)}$), we provide the performance of two-layer Warpformer on MIMIC-III-based datasets (Figure~\ref{fig:sensitive1d}) and the AUROC of a four-layer Warpformer on the PhysioNet dataset (Figure~\ref{fig:physio3d_auroc}). It shows that MIMIC-III-based datasets consistently yield the best results when $\tilde{L}^{(1)}=1$. Additionally, the results of the four-scale experiment on the PhysioNet dataset do not surpass the performance of the 3-scale settings. Notably, extreme downsampling applied to $\tilde{L}^{(1)}$ and $\tilde{L}^{(2)}$ achieves relatively strong performance, aligning with our observations in Section~\ref{sec:sensitive}.